\documentclass[11pt]{article}

\usepackage[final]{acl}

\usepackage{times}
\usepackage{latexsym}

\usepackage[T1]{fontenc}
\usepackage[utf8]{inputenc}
\DeclareUnicodeCharacter{201C}{``}
\DeclareUnicodeCharacter{201D}{''}
\DeclareUnicodeCharacter{2014}{---}

\usepackage{microtype}

\usepackage{inconsolata}

\usepackage{graphicx}
\usepackage{subcaption}
\usepackage{booktabs}
\usepackage{amsmath}
\usepackage{algorithm}
\usepackage{algpseudocode}
\usepackage{tcolorbox}
\usepackage{enumitem}
\usepackage[misc]{ifsym}
\algrenewcommand\algorithmicrequire{\textbf{Require:}}
\algrenewcommand\algorithmicensure{\textbf{Ensure:}}
\usepackage{amsmath}

\usepackage{multirow}

\title{\texorpdfstring{\raisebox{-0.8ex}{\includegraphics[height=1.8em]{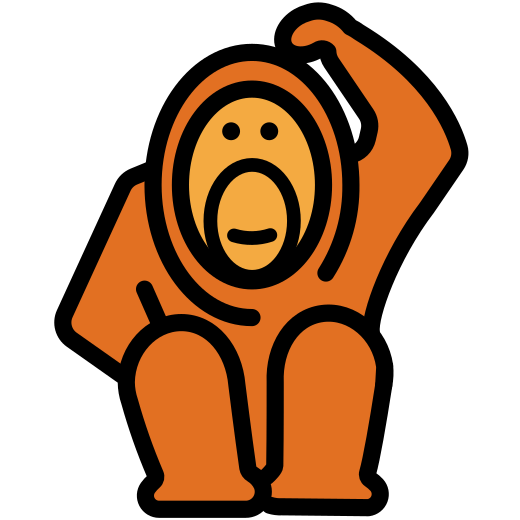}}}{} Thinking with Reasoning Skills: Fewer Tokens, More Accuracy}

\author{
 \textbf{Guangxiang Zhao\textsuperscript{1,}\footnotemark[5]},
 \textbf{Qilong Shi\textsuperscript{2,}\footnotemark[5]},
 \textbf{Xusen Xiao\textsuperscript{3}},
 \textbf{Xiangzheng Zhang\textsuperscript{1,}\textsuperscript{\Letter}}
 \textbf{Tong Yang\textsuperscript{4}},
 \textbf{Lin Sun\textsuperscript{1,}\textsuperscript{\Letter}}
\\
 \textsuperscript{1}Qiyuan Tech,
 \textsuperscript{2}Tsinghua University,
 \textsuperscript{3}The University of Hong Kong,
 \textsuperscript{4}Peking University
\\
\texttt{zhaoguangxiang@pku.edu.cn},
\texttt{stallone@pku.edu.cn},
\texttt{xiaoxusen@connect.hku.hk},\\
\texttt{zhangxiangzheng@360.cn},
\texttt{yangtong@pku.edu.cn},
\texttt{sunlin1@360.cn}
\\
 \small{
   \Letter\ \textbf{Correspondence:} \href{mailto:zhangxiangzheng@360.cn}{zhangxiangzheng@360.cn},
   \href{mailto:sunlin1@360.cn}{sunlin1@360.cn}
 }
}

\begin{document}
\renewcommand{\thefootnote}{\fnsymbol{footnote}}
\maketitle
\footnotetext[1]{Code: \url{https://github.com/stallone0000/Reasoning-Skill}. Dataset: \url{https://huggingface.co/datasets/stallone0000/Reasoning-Skill}. Demo: \url{https://reasoning-skill.onrender.com/}. The demo provides side-by-side comparisons of Direct and TRS reasoning. Latest demo URL will be updated in GitHub.}
\footnotetext[5]{The first two authors contributed equally.}
\begin{abstract}
Reasoning LLMs often spend substantial tokens on long intermediate reasoning traces (e.g., chain-of-thought) when solving new problems. We propose to summarize and store reusable reasoning skills distilled from extensive deliberation and trial-and-error exploration, and to retrieve these skills at inference time to guide future reasoning. Unlike the prevailing \emph{reasoning from scratch} paradigm, our approach first recalls relevant skills for each query, helping the model avoid redundant detours and focus on effective solution paths. We evaluate our method on coding and mathematical reasoning tasks, and find that it significantly reduces reasoning tokens while improving overall performance. The resulting lower per-request cost indicates strong practical and economic potential for real-world deployment.
\end{abstract}

\section{Introduction}

Reasoning-centric large language models (LRMs) have rapidly evolved from research novelties to standard capabilities. Modern models, such as OpenAI o1 and DeepSeek-R1, now explicitly encourage intermediate deliberation to enhance reliability in math and code \cite{openai2024openaio1card,guo2025deepseek,google_gemini3_2025, anthropic_claude_opus_46_2026,singh2025openaigpt5card}. While this shift delivers impressive accuracy, it introduces a major bottleneck: \emph{test-time compute is paid in tokens and latency}.

In practice, LRMs often generate thousands of ``thinking'' tokens, comprising redundant verification and trial-and-error loops \cite{han2024tale,wang2025nowait}. Since commercial APIs bill by token—often pricing output higher than input \cite{openai_pricing}—lengthy traces dominate query costs and latency. Industry reports confirm that reasoning-heavy inference significantly amplifies infrastructure strain \cite{uptime_reasoning_footprint}. Consequently, efficient reasoning is production-critical: we seek the benefits of deliberation without the cost of repeated rediscovery.

\begin{figure}[t]
  \centering
    \includegraphics[width=\linewidth]{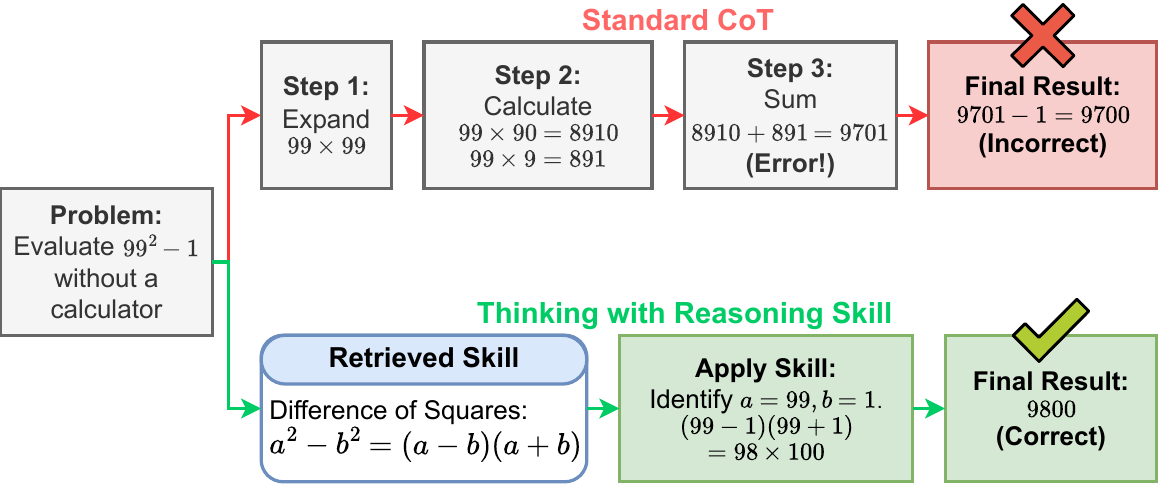}

    \vspace{0.5cm}
    \includegraphics[width=\linewidth]{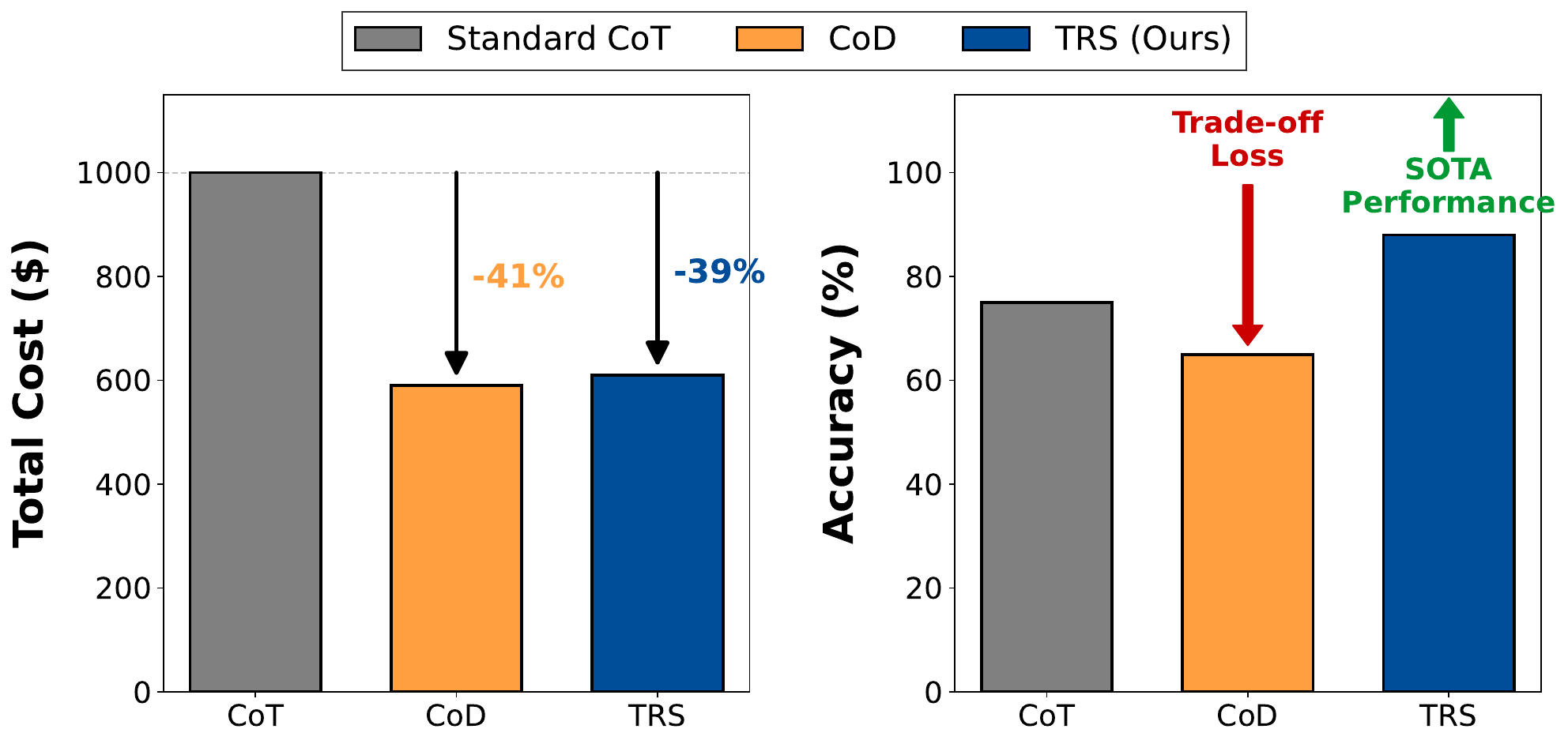}
  \caption{\textbf{Above:} The "Gist" of Thinking with Reasoning Skills. \textbf{Below:} Breaking the Efficiency-Accuracy Trade-off.}
  \label{fig:intuition2}
  \vspace{-0.3cm}
\end{figure}

Recent work has proposed to shorten reasoning traces by controlling \emph{how} models think.
Prompt-only approaches compress intermediate traces (e.g., Chain-of-Draft) \cite{xu2025cod} or enforce explicit token budgets (e.g., TALE) \cite{han2024tale};
decoding-time approaches suppress reflection markers to avoid overthinking loops (e.g., NoWait) \cite{wang2025nowait}.
However, these methods still treat each query as a blank slate: the model is asked to re-derive solution logic from scratch, only faster.
This often yields an efficiency--accuracy trade-off on hard problems: when the reasoning space is forcibly compressed, models may skip crucial steps and fail.

Humans solve this differently.
Experts rarely re-derive everything; they recall reusable \emph{skills} distilled from past problem solving (e.g., ``look for an invariant'', ``use two-pointers'', ``apply chain rule'').
Notably, \emph{skill libraries and experience memory have become a mainstream design pattern in LLM agents}, enabling systems to reuse prior reflections or executable skills across tasks without weight updates \cite{yao2022react,shinn2023reflexion,wang2023voyager}.
Yet, for \emph{pure reasoning} (math/coding) under token budgets, we still largely operate in a ``reasoning from scratch'' regime.

We propose \textbf{Thinking with Reasoning Skills (TRS)} (Figure~\ref{fig:intuition2}): a training-free, retrieval-augmented framework that decouples \emph{acquiring} reasoning logic from \emph{executing} reasoning.
Offline, we distill long deliberation trajectories (including trial-and-error) into compact, reusable \emph{reasoning skills} with explicit triggers and pitfalls.
Online, we retrieve the most relevant skills for a new query (even when no near-duplicate problems exist) and inject them into the prompt to steer the model toward an effective solution path.
The key idea is that while questions are open-ended, \emph{reasoning patterns are reusable}; we can replace redundant internal rediscovery with externalized procedural memory.

Empirically, TRS yields a surprising result: across both mathematical and coding benchmarks, it \emph{reduces thinking tokens} while \emph{improving accuracy} compared to standard reasoning from scratch, especially on harder subsets and for weaker models.
This indicates that efficiency need not come at the expense of correctness---if we reuse distilled experience rather than merely forcing brevity.

Our main contributions are:
\begin{itemize}
  \item \textbf{A practical framework for token-efficient reasoning:} TRS distills deliberation into retrievable \emph{reasoning skills} and reuses them at inference time, remaining compatible with black-box LLM APIs.
  \item \textbf{Breaking the efficiency--accuracy trade-off:} on math and coding tasks, TRS consistently reduces thinking tokens/cost while matching or improving accuracy over strong baselines (e.g., CoD/TALE-style prompts).
  \item \textbf{Analysis for deployment:} we study transferability across models, output length, and retrieval ablations, showing TRS benefits grow on harder problems and can transfer across heterogeneous LLMs.
\end{itemize}

\begin{figure*}[htbp]
  \centering
    \includegraphics[width=\linewidth]{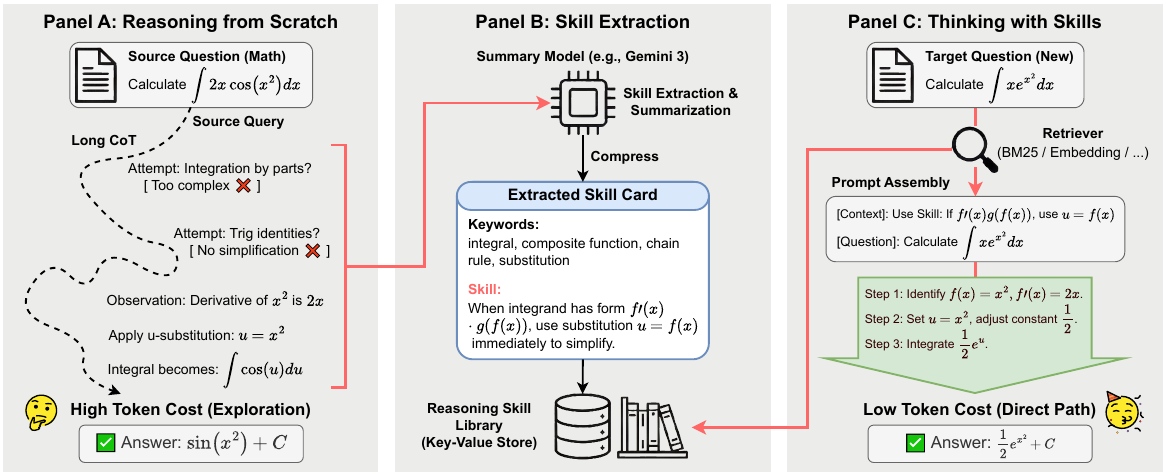}
  \caption{The process of Thinking with Reasoning Skills (TRS).
\textbf{(A)} A standard reasoning model solves a source problem via a "reasoning from scratch" approach, involving redundant steps and trial-and-error (high token cost).
\textbf{(B)} A summarizer model distills this long trajectory into a reusable, abstract reasoning skill (e.g., identifying the Chain Rule pattern), which is stored in a Key-Value library.
\textbf{(C)} When a new, independent problem with similar underlying logic is presented, the relevant skill is retrieved and injected into the prompt. This guides the model to follow a direct solution path, significantly reducing reasoning tokens while maintaining high accuracy.}
  \label{fig:framework}
  \vspace{-0.5cm}
\end{figure*}

\section{Related Work}

\paragraph{Deliberative reasoning and test-time compute.}
Chain-of-Thought (CoT) prompting improves multi-step reasoning by eliciting intermediate traces \cite{wei2022cot}, and subsequent work amplifies this effect via sampling/aggregation (self-consistency) \cite{wang2022selfconsistency} or structured search over thoughts (e.g., Tree-of-Thoughts) \cite{yao2023tot}.
Agent-style prompting (e.g., ReAct) further couples reasoning with actions/tools \cite{yao2022react}.
While powerful, these paradigms often increase test-time compute, motivating the recent surge of \emph{efficient reasoning} research.

\paragraph{Token-efficient reasoning via compression, budgets, decoding, or post-training.}
A first line of work reduces output length purely by prompting, e.g., compressing traces into minimalist drafts (Chain-of-Draft) \cite{xu2025cod}, or explicitly constraining reasoning budgets (TALE) \cite{han2024tale}.
Decoding-time interventions suppress reflection markers correlated with redundant overthinking loops (NoWait) \cite{wang2025nowait}, but typically require logit-level control.
Other approaches modify the model to internalize efficiency (e.g., post-training for budget awareness \cite{han2024tale}) or shift reasoning into alternative representations (e.g., latent-space reasoning in Coconut) \cite{hao2024coconut}.
In parallel, recent work explores pruning/sparsifying reasoning computation for lower cost (e.g., reasoning-aware attention sparsity) \cite{hu-etal-2025-reasoning-aware}.
TRS is complementary: instead of only ``thinking shorter'' or changing decoding/training, we reduce detours by reusing distilled procedural experience.

\paragraph{Retrieval, memory, and reusable skills / templates.}
Retrieval augmentation has been studied beyond factual QA, showing gains even for non-knowledge-intensive tasks when retrieval is integrated properly \cite{guo2023pgra,chen2022rapl}.
In agent systems, memory and skill libraries reuse past experience to improve future behavior without weight updates (e.g., Reflexion) \cite{shinn2023reflexion}, and embodied agents can accumulate an explicit skill library for compositional reuse (e.g., Voyager) \cite{wang2023voyager}.
Closest to our framing, recent ``thought template'' methods build reusable reasoning templates (BoT) \cite{yang2024bot} or distill templates into smaller models with self-correction (SuperCorrect) \cite{yang2025supercorrect}, and caching frameworks retrieve prior high-quality reasoning outputs to assist cheaper models (Cache-of-Thought) \cite{deng-etal-2025-cache-of-thought}.
TRS differs in goal and design: we target \emph{token-efficient reasoning} for math/coding in a black-box-compatible pipeline, distilling both \emph{successful strategies and failure-mode fixes} into compact \emph{reasoning skills}, and we systematically study cross-model transfer, gating strength, skill length, and retrieval ablations under difficulty slicing.

\section{Methodology}
\label{sec:method}

\paragraph{Goal and interface.}
We target \emph{production-style} reasoning where cost and latency scale with test-time tokens.
Given a query $q$ (math or coding), a reasoning model $\mathcal{M}_r$ generates an intermediate trace $\tau$ and a final output $y$.
Correctness is determined by $\mathrm{Eval}(y,a)$: exact match for math, and pass@1 against unit tests for code.
Our goal is to \emph{reduce reasoning tokens} (thinking length) while maintaining or improving accuracy. Unless the model exposes separate thinking tokens, we use completion tokens as a consistent proxy across methods.

\subsection{Thinking with Reasoning Skills (TRS)}
Figure~\ref{fig:framework} illustrates TRS.
The key idea is to replace repeated ``reasoning from scratch'' with \emph{retrieved procedural experience}.
TRS is \textbf{training-free} and \textbf{black-box compatible}:
offline we distill long trajectories into reusable skill cards;
online we retrieve and inject a few cards to guide $\mathcal{M}_r$ toward a direct solution path.
This decouples expensive exploration (one-time) from cheap reuse (per request).

\subsection{Offline Skill Library Construction}
\label{sec:skill-extraction}
For each source instance $(q_i,a_i)$, we run $\mathcal{M}_r$ to obtain $(\tau_i,y_i)$ and compute $c_i=\mathrm{Eval}(y_i,a_i)$.
We form an \emph{experience record} $(q_i,a_i,\tau_i,y_i,c_i)$ and distill it with a stronger summarizer $\mathcal{M}_s$ into:
(i) a compact \emph{skill card} $s_i$ and (ii) retrieval triggers $K_i$ (10--20 keywords).
If $c_i{=}1$, $s_i$ captures the essential pattern (minimal transformation / invariant / algorithmic template).
If $c_i{=}0$, $s_i$ captures a reusable \emph{failure-mode fix} (anti-pattern $\rightarrow$ correction).
We standardize skill cards as short structured text:
\textbf{Trigger / Do / Avoid / Check / Risk}.
Extraction prompts, schema constraints, and card validation rules are in Appendix~\ref{app:skillformat}.

\paragraph{Key--value storage.}
Each card becomes a key--value entry in a library $\mathcal{L}=\{(x_i \rightarrow v_i)\}$,
where $v_i=s_i$ and $x_i=\mathrm{Concat}(q_i,K_i)$.
Concatenating $q_i$ with trigger keywords improves recall for sparse retrieval while keeping keys human-interpretable.

\subsection{Online Retrieval and Skill Injection}
\label{sec:retrieval-inference}
Given a new query $q$, TRS retrieves top-$k$ relevant skill cards and prepends them to the prompt of $\mathcal{M}_r$.

\paragraph{Retrieval backends.}
We support sparse BM25, dense retrieval with embeddings + nearest-neighbor search, and hybrid retrieval.
We use \textbf{BM25} as default for math and \textbf{hybrid} for coding, following ablations (Appendix~\ref{app:retrieval}).

\paragraph{Prompting and lightweight gating.}
Retrieved skills may be partially irrelevant or conflicting.
We therefore use a lightweight arbitration instruction:
\emph{use only directly applicable skills; ignore irrelevant/contradictory advice}.
To prevent prompt inflation, we (i) cap each card to a small budget (few bullets), and (ii) truncate to top-$k$ by retrieval score.
We analyze gating strength and skill length budgets in Section~\ref{sec:ablation}.

\begin{figure}[htbp]
    \centering
    \begin{tcolorbox}[
      title={TRS prompt template},
      colback=gray!3,
      colframe=black!35,
      boxrule=0.6pt,
      arc=1.5mm,
      left=1.5mm,right=1.5mm,top=1mm,bottom=1mm
    ]
    \small
    \textbf{Retrieved Reasoning Skill(s).}
    \begin{enumerate}[leftmargin=1.2em,itemsep=0.15em]
      \item \textbf{Skill $s_{(1)}$.} [A short, actionable skill distilled from prior trajectories.]
      \item \textbf{Skill $s_{(2)}$.} [Optional; only if $k>1$.]
      \item \dots
    \end{enumerate}

    \vspace{0.25em}
    \textbf{Instruction.}
    Prefer the most directly applicable skill; ignore irrelevant or contradictory advice.
    Keep intermediate reasoning concise while maintaining correctness.

    \vspace{0.25em}
    \textbf{Task.}
    Solve the following question. Use the retrieved skills above when relevant.

    \vspace{0.25em}
    \textbf{Question.}
    [Insert the new query $q$ here.]
    \end{tcolorbox}

    \caption{A standard TRS prompt template that injects retrieved reasoning skills before the query.}
    \label{fig:trs-template}
    \vspace{-0.5cm}
\end{figure}

\paragraph{Why this reduces tokens in practice.}
TRS reduces reasoning tokens by replacing redundant exploration (branching, detours, repeated debugging) with a retrieved procedural shortcut and explicit pitfalls distilled from failures.
Although TRS adds a small input prefix, it typically yields a larger reduction in generated reasoning tokens, leading to lower end-to-end cost and latency; we quantify token/cost trade-offs in our experiments.

\begin{table*}[t]
\centering
\small
\begin{tabular}{l|l|lll|lll}
\toprule
\textbf{Model} & \textbf{Method} & \multicolumn{3}{c|}{\textbf{Math}} & \multicolumn{3}{c}{\textbf{Coding}} \\
 &  & \textbf{Acc.} & \textbf{Token \#} & \textbf{Cost} & \textbf{Acc.} & \textbf{Token \#} & \textbf{Cost} \\
\midrule
\multirow{2}{*}{Gemini-3-Flash} & Direct & 94.8\% & 7490 & 100.0\% & 72.0\% & 20,206 & 100.0\% \\
 & TRS & 95.5\% \textcolor{cyan}{$_{\uparrow 0.7\%}$} 
 & 6106 \textcolor{cyan}{$_{\downarrow 18.5\%}$} 
 & 82.5\% \textcolor{cyan}{$_{\downarrow 17.5\%}$}
 & 71.7\% \textcolor{red}{$_{\downarrow 0.3\%}$} 
 & 17,072 \textcolor{cyan}{$_{\downarrow 15.5\%}$}
 & 85.2\% \textcolor{cyan}{$_{\downarrow 14.8\%}$} \\
\midrule
\multirow{2}{*}{Doubao Seed\textsuperscript{$\dagger$}} & Direct & 95.1\% & 3453 & 100.0\% & 63.6\% & 7,500 & 100.0\% \\
 & TRS & 94.9\% \textcolor{red}{$_{\downarrow 0.2\%}$} 
 & 1411 \textcolor{cyan}{$_{\downarrow 59.1\%}$}
 & 46.2\% \textcolor{cyan}{$_{\downarrow 53.8\%}$}
 & 64.4\% \textcolor{cyan}{$_{\uparrow 0.8\%}$} 
 & 6,730 \textcolor{cyan}{$_{\downarrow 10.3\%}$}
 & 94.0\% \textcolor{cyan}{$_{\downarrow 6.0\%}$} \\
\midrule
\multirow{2}{*}{GPT-OSS-120B} & Direct & 93.7\% & 1257 & 100.0\% & 54.2\% & 5,080 & 100.0\% \\
 & TRS & 93.7\% \textcolor{gray}{$_{\rightarrow 0\%}$} 
 & 976 \textcolor{cyan}{$_{\downarrow 22.4\%}$}
 & 83.1\% \textcolor{cyan}{$_{\downarrow 16.9\%}$}
 & 58.3\% \textcolor{cyan}{$_{\uparrow 4.1\%}$} 
 & 5,177 \textcolor{red}{$_{\uparrow 1.9\%}$}
 & 104.8\% \textcolor{red}{$_{\uparrow 4.8\%}$} \\
% \midrule
% \multirow{2}{*}{DeepSeek-R1} & CoT & -- & -- & -- & -- & -- & -- \\
%  & Ours & -- & -- & -- & -- & -- & -- \\
\midrule
\multirow{2}{*}{GPT-4o-mini} & Direct & 59.6\% & 819 & 100.0\% & 22.0\% & 726 & 100.0\% \\
 & TRS & 61.4\% \textcolor{cyan}{$_{\uparrow 1.8\%}$} 
 & 650 \textcolor{cyan}{$_{\downarrow 20.6\%}$}
 & 93.1\% \textcolor{cyan}{$_{\downarrow 6.9\%}$}
 & 24.4\% \textcolor{cyan}{$_{\uparrow 2.4\%}$} 
 & 480 \textcolor{cyan}{$_{\downarrow 33.9\%}$}
 & 93.7\% \textcolor{cyan}{$_{\downarrow 6.3\%}$} \\
\bottomrule
\end{tabular}
\caption{Comparison of \textsc{Direct} and TRS on math and coding datasets. Token \# denotes average \emph{completion} tokens. Cost is normalized to Direct = 100.0\% within each model--dataset pair under a fixed input/output token price ratio.
\textsuperscript{$\dagger$}We use different Doubao Seed versions: Seed-1.8 Pro for math and Seed-2.0 Pro for coding. Cyan indicates improvement.}
\label{tab:main}
\vspace{-0.5cm}
\end{table*}

\section{Experimental Setup}
\label{sec:exp-setup}

\paragraph{Math benchmark (disjoint split).}
We evaluate on \textsc{DeepMath-103K} with a strict split:
\textbf{93K} source problems for library construction and \textbf{10K} held-out problems for evaluation.
Unless stated otherwise, we use BM25 retrieval with $k{=}1$ on math, which is consistently best in our analysis (Appendix~\ref{app:retrieval}).

\paragraph{Coding benchmark (end-to-end judging).}
We evaluate on \textsc{Nemotron-Competitive-Programming-v1} (34K).
We filter to \textbf{26.6K} instances with reliable local judging signals and hold out \textbf{1K} for testing.
We generate solution trajectories and verdicts using two generators (Gemini 3 Flash Preview and Doubao Seed 2.0 Pro), distill structured experience cards with Gemini 3 Flash Preview, and build a dual-route retrieval index (BM25 + dense).
At inference time, we retrieve \textbf{top-5} cards and inject them into the prompt; the model receives only (cards + problem) with no additional reasoning-style prompts.
Final correctness is determined by a local compile-and-run judge on test cases (pass@1).
Full pipeline, filtering rules, outcome partitions, and card statistics are in Appendix~\ref{app:coding} and Appendix~\ref{app:judge}.

\paragraph{Models and summarizer.}
For inference-time evaluation, we compare TRS against strong baselines on multiple models, including \textbf{GPT-OSS-120B}, \textbf{Gemini-3 Flash Preview}, and \textbf{GPT-4o-mini}. We also evaluate on the \textbf{Doubao Seed} family: we use \textbf{Seed-1.8 Pro} for math and \textbf{Seed-2.0 Pro} for coding, since the coding experiments were conducted later and thus adopted the latest available Seed model.
For skill extraction across all experiments, we use a fixed strong summarizer (Gemini Flash) and fixed distillation prompts (Appendix~\ref{app:skillformat}).
For dense retrieval, we use BGE-M3 embeddings with FAISS (Appendix~\ref{app:retrieval}).

\subsection{Metrics}
\label{sec:exp-metrics}
We report: \textbf{Accuracy} (exact match for math; pass@1 for coding), \textbf{Reasoning tokens} (think tokens when exposed; otherwise output length proxy), and \textbf{Cost reduction} relative to the baseline under a representative input/output token pricing ratio.
Token accounting details are provided in Appendix~\ref{app:cost}.

\subsection{Difficulty Slices by Baseline Thinking Length}
\label{sec:exp-threshold}
To isolate hard instances, we slice the test set by the baseline model's thinking length.
Let $T_{\text{think}}^{\text{base}}(q)$ be baseline think tokens for query $q$.
For threshold $\theta$, define $\mathcal{Q}_{\ge \theta}=\{q: T_{\text{think}}^{\text{base}}(q)\ge\theta\}$.
Larger $\theta$ indicates harder instances (baseline ``thinks longer'').
We report accuracy and token/cost trends across thresholds to quantify where TRS yields the largest benefits.

\section{Main Results}
\label{sec:main-results}
Table~\ref{tab:main} shows that TRS improves the accuracy--efficiency trade-off on both math and coding.
On \textbf{math}, TRS reduces completion tokens and cost while preserving or improving accuracy: Gemini-3-Flash improves by +0.7 with a 17.5\% cost reduction, GPT-4o-mini improves by +1.8 with lower cost, Doubao Seed cuts cost by 53.8\% with only a $-0.2$ change, and GPT-OSS-120B preserves accuracy while reducing cost by 16.9\%.
On \textbf{coding}, TRS generally improves pass@1 with lower cost when measured: GPT-4o-mini increases to 24.4\% (+2.4) with a 6.3\% cost reduction, Doubao Seed improves by +0.8 with a 6.0\% cost reduction, and GPT-OSS-120B improves by +4.1 though cost slightly increases (+4.8\%) due to larger prompts.
Overall, TRS gains efficiency by \emph{reusing distilled procedural skills} instead of forcing shorter reasoning, often reducing per-request cost without sacrificing correctness.

\section{Analysis}
\label{sec:ablation}

\subsection{Comparison with Baselines}
\label{sec:main-comparison}
\begin{figure}[tbp]
%\vspace{-0.1cm}
%\setlength{\abovecaptionskip}{-0cm}
  \centering
  \includegraphics[width=\linewidth]{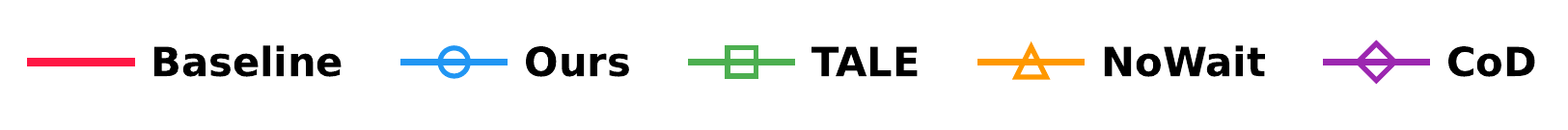}
  \begin{minipage}[t]{0.49\linewidth}
    \centering
    \includegraphics[width=\linewidth]{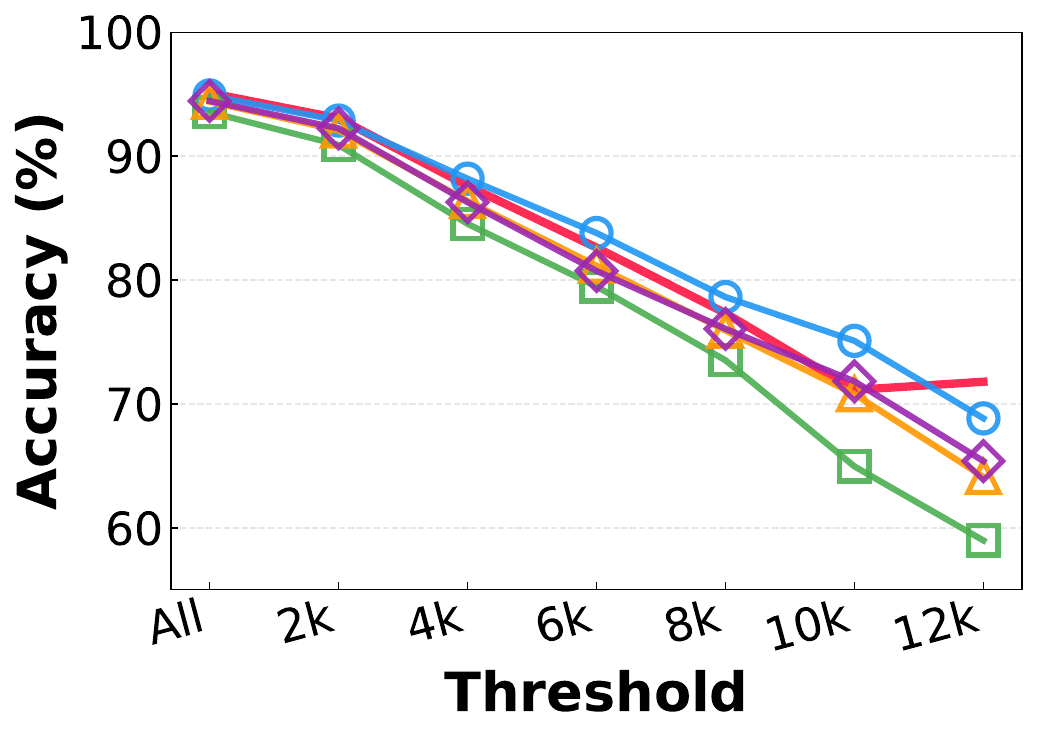}
    \subcaption{Doubao: Accuracy (\%)}
  \end{minipage}\hfill
  \begin{minipage}[t]{0.49\linewidth}
    \centering
    \includegraphics[width=\linewidth]{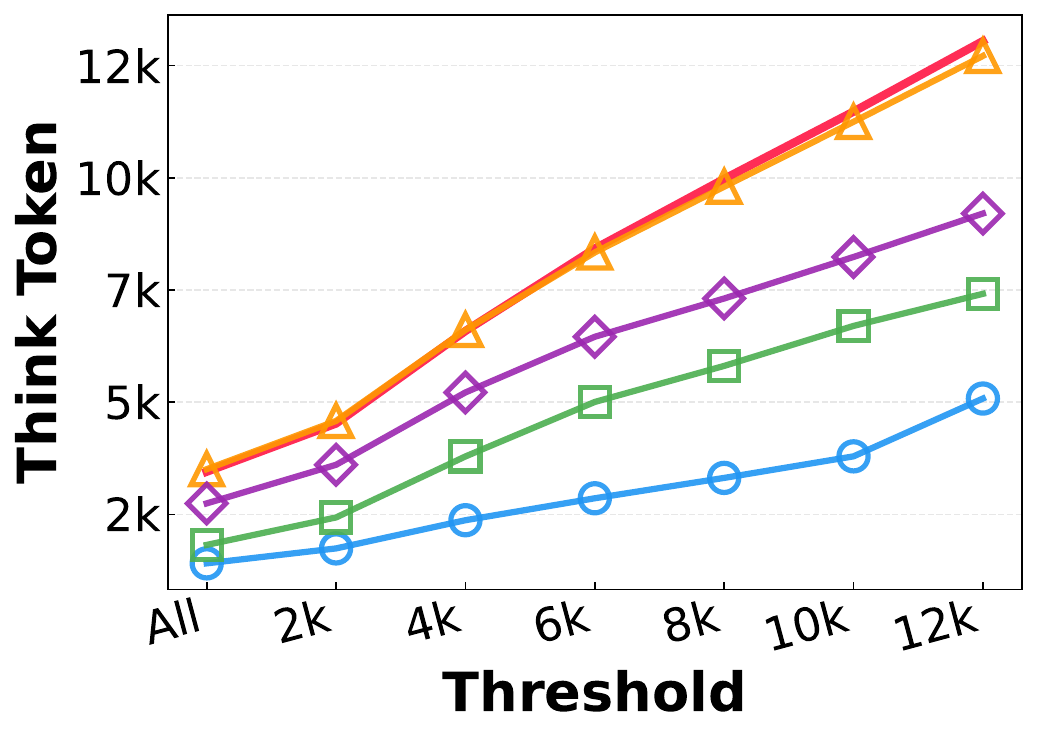}
    \subcaption{Doubao: Think Tokens}
  \end{minipage}

  \begin{minipage}[t]{0.49\linewidth}
    \centering
    \includegraphics[width=\linewidth]{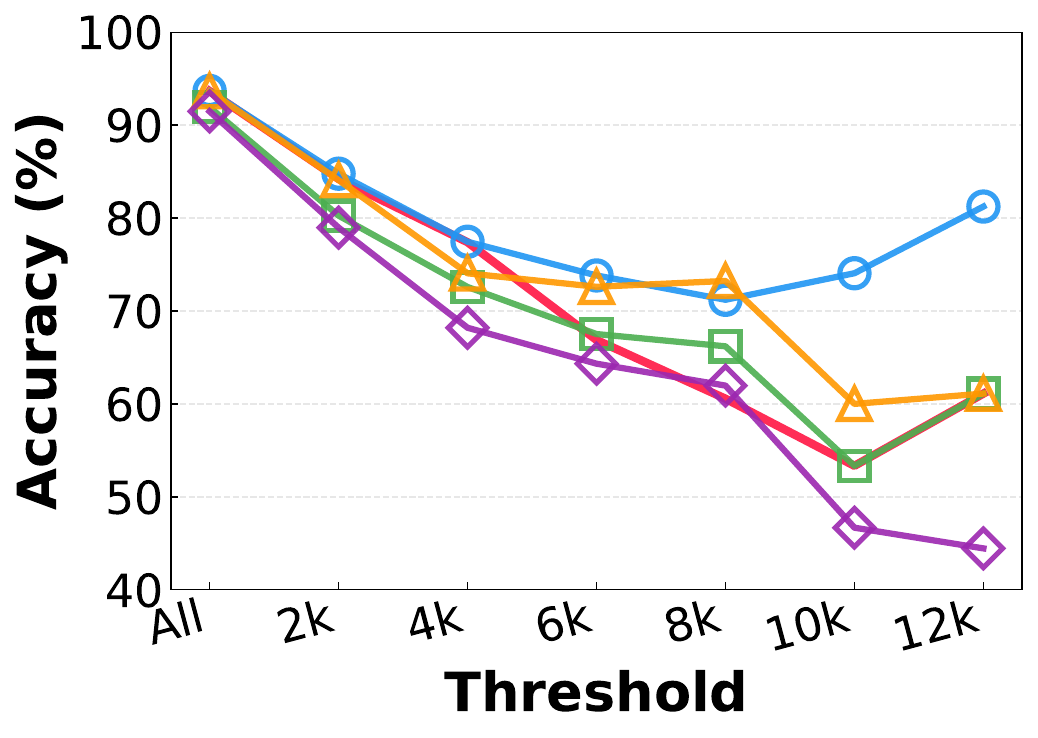}
    \subcaption{OSS: Accuracy (\%)}
  \end{minipage}\hfill
  \begin{minipage}[t]{0.49\linewidth}
    \centering
    \includegraphics[width=\linewidth]{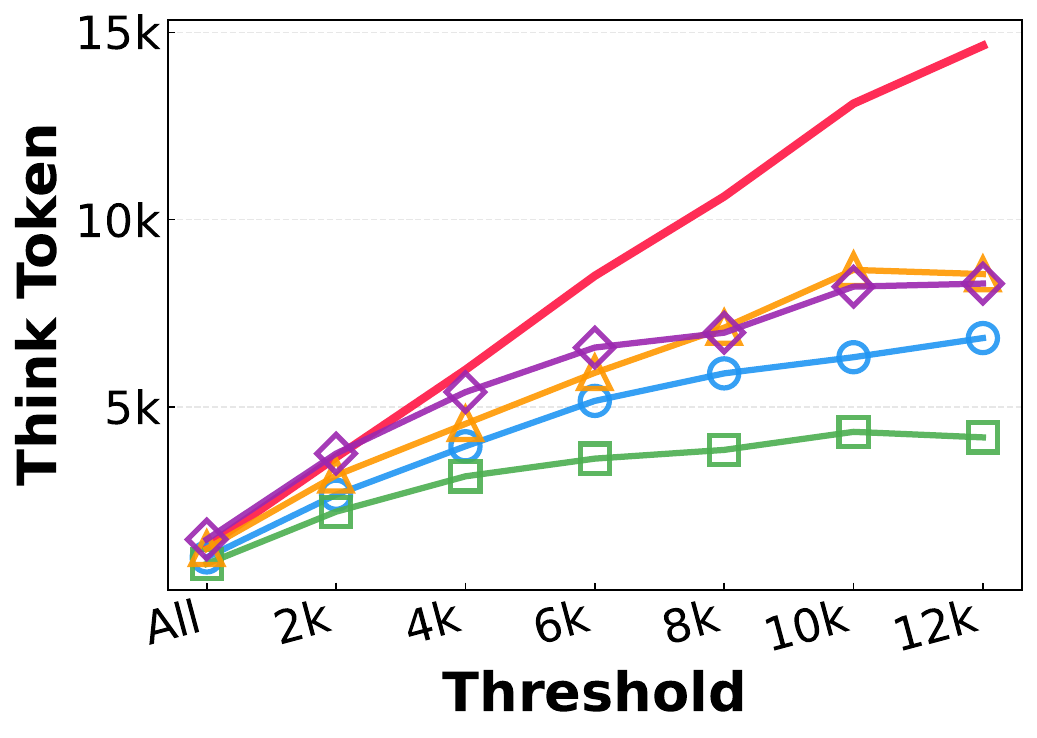}
    \subcaption{OSS: Think Tokens}
  \end{minipage}
  \caption{Performance comparison across difficulty thresholds (Doubao vs. OSS).}
  \label{fig:compare-main-6}
  %\vspace{-0.5cm}
\end{figure}

We compare \textbf{TRS} against \textbf{TALE-EP} (budget-constrained), \textbf{Chain of Draft (CoD)} (brevity-constrained), and \textbf{NoWait}. Figure~\ref{fig:compare-main-6} details performance sliced by baseline think token count $\theta$.

\begin{figure}[htbp]
%\vspace{-0.1cm}
%\setlength{\abovecaptionskip}{-0cm}
  \centering
  \includegraphics[width=\linewidth]{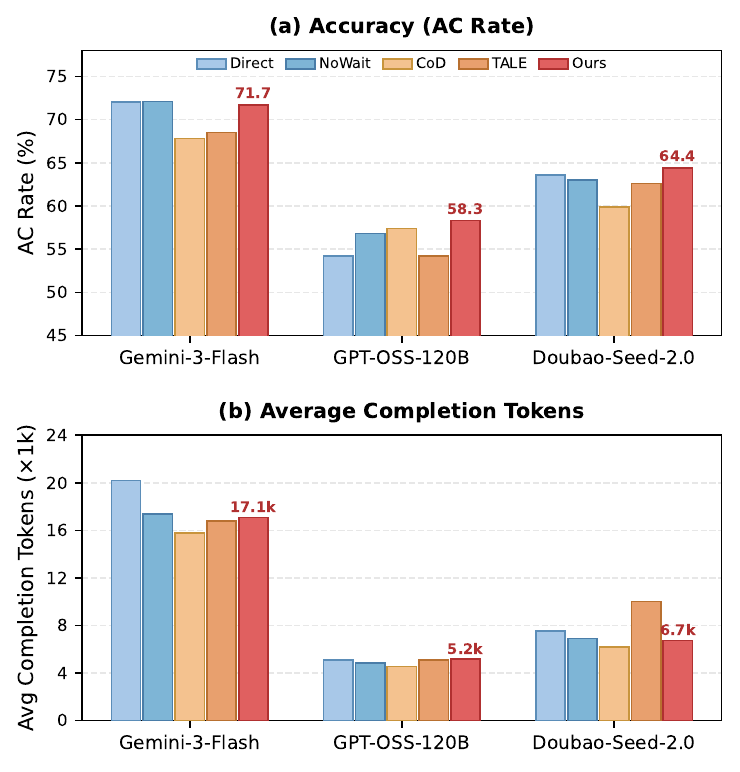}
  \caption{Compare to Direct on coding competitions.}
  %\vspace{-0.5cm}
\end{figure}

\paragraph{Breaking the Trade-off.} Existing methods suffer severe accuracy trade-offs on hard problems. As difficulty $\theta$ increases, TALE and CoD show catastrophic collapse (e.g., dropping below baseline on GPT-OSS for $\theta > 10$k), suggesting enforced brevity cripples deep reasoning. In contrast, \textbf{TRS} consistently matches or exceeds baseline accuracy, achieving significant uplift ($\sim$45\% to $\sim$80\%) on the hardest GPT-OSS tasks. Retrieved skills act as a ``navigation map,'' preventing the model from getting lost in incorrect branches.

\paragraph{Efficiency Gains.} On Doubao, TRS achieves the \textbf{lowest} token consumption, cutting generated tokens for hard problems ($\theta \ge 12$k) from $\sim$12k to $\sim$5k while maintaining top accuracy. On GPT-OSS, TRS strikes the optimal balance: it halves token usage ($\sim$15k to $\sim$7k) without the accuracy loss seen in ``speed-limit'' approaches like TALE or CoD.

\subsection{Controlled Representation and Coverage Analysis}
\label{sec:controlled-representation}
To isolate the effect of \emph{structured reasoning-skill distillation} from simple retrieval, we ran a same-setup control on a fixed 500-question DeepMath subset, keeping the evaluation subset, target model, prompt family, verifier, retriever, and top-$k$ fixed while varying only the retrieved representation. Table~\ref{tab:repr-coverage} shows that naive retrieval does not explain the TRS gains: raw examples, raw CoT traces, free-form summaries, and small structured-card libraries all underperform full TRS, especially on OSS. The strongest result appears only when structured skill reuse is paired with sufficient library coverage, indicating that TRS is not merely prepending relevant-looking context but retrieving reusable procedural guidance from a broad skill bank.

\begin{table}[t]
    \centering
    \small
    \setlength{\tabcolsep}{3.3pt}
    \begin{tabular}{llcc}
        \toprule
        Analysis & Condition & Doubao & OSS \\
        \midrule
        \multirow{6}{*}{Representation}
        & Direct & 92.8 & 90.6 \\
        & Raw examples & 92.0 & 87.2 \\
        & Raw CoT & 91.0 & 89.8 \\
        & Free summary & 91.6 & 86.2 \\
        & Structured card (2.5k) & 91.2 & 85.2 \\
        & Full TRS (93k) & 92.6 & 91.2 \\
        \midrule
        \multirow{4}{*}{Coverage}
        & 10k structured cards & 91.8 & 89.2 \\
        & 30k structured cards & 90.6 & 87.4 \\
        & 60k structured cards & 92.2 & 90.8 \\
        & 93k structured cards & 92.6 & 91.2 \\
        \bottomrule
    \end{tabular}
    \caption{Controlled representation and coverage analysis on a fixed 500-question DeepMath subset. Values are accuracy percentages; only the retrieved representation or structured-library size changes.}
    \label{tab:repr-coverage}
\end{table}

\subsection{Transfer to External Contest-Math Benchmarks with AoPS-Derived Skills}
\label{sec:aops-benchmark-transfer}
To test whether skills distilled from a separate contest-math corpus transfer beyond DeepMath, we built an external AoPS-derived library from 7,616 contest-math problems and evaluated TRS on 120 benchmark questions spanning AIME 2024 I, AIME 2024 II, AIME 2025, AIME 2026, and HMMT November 2025. In this study, direct prompting uses 32 repeats per question and TRS uses 8 repeats per question; TRS retrieves the top-1 BM25 match from the AoPS library and injects only the retrieved heuristic into a unified prompt. Table~\ref{tab:aops-benchmark-summary} shows a heterogeneous but meaningful transfer pattern: 13 of 25 model-benchmark pairs improve accuracy and 20 of 25 reduce per-query cost. The clearest average beneficiary is Doubao-1.8 (+1.88 accuracy points with a 2.8\% cost reduction), while GPT-OSS-120B shows smaller but still positive average gains (+0.50) with a 6.5\% cost reduction. By contrast, Gemini-3-Flash improves accuracy (+0.92) but increases output tokens and cost, and Doubao-2.0-Pro mainly trades a small amount of accuracy for a 15.7\% cost reduction. We therefore treat external skill transfer as evidence of reusable reasoning structure in nearby domains, but not as a universal improvement guarantee. Appendix~\ref{app:aops-benchmark} provides the full 25-condition accuracy and cost-percentage heatmaps and setup details.

\begin{table}[t]
    \centering
    \small
    \setlength{\tabcolsep}{3.5pt}
    \begin{tabular}{lccc}
        \toprule
        Model & Acc. $\Delta$ & Out Tok. $\Delta$ & Cost $\Delta$ \\
        \midrule
        Doubao-1.8 & \textcolor{cyan}{+1.88} & \textcolor{cyan}{$\downarrow$4.9\%} & \textcolor{cyan}{$\downarrow$2.8\%} \\
        Doubao-2.0-Pro & \textcolor{red}{-1.11} & \textcolor{cyan}{$\downarrow$17.5\%} & \textcolor{cyan}{$\downarrow$15.7\%} \\
        Gemini-3-Flash & \textcolor{cyan}{+0.92} & \textcolor{red}{$\uparrow$8.6\%} & \textcolor{red}{$\uparrow$9.0\%} \\
        GPT-OSS-120B & \textcolor{cyan}{+0.50} & \textcolor{cyan}{$\downarrow$8.1\%} & \textcolor{cyan}{$\downarrow$6.5\%} \\
        GPT-OSS-20B & \textcolor{red}{-0.60} & \textcolor{cyan}{$\downarrow$5.3\%} & \textcolor{cyan}{$\downarrow$4.3\%} \\
        \bottomrule
    \end{tabular}
    \caption{Average TRS-minus-direct deltas on the 120-question external contest-math suite using the pure 7,616-card AoPS library. Accuracy is reported in percentage points; output-token and cost changes are relative to Direct, following the accounting style of Table~\ref{tab:main}. Because direct and TRS use 32 and 8 repeats per question, respectively, we treat this comparison as descriptive rather than a formal significance test.}
    \label{tab:aops-benchmark-summary}
\end{table}

\subsection{Cross-Model Skill Transfer}
\label{sec:ablation-cross}
\begin{figure}[t]
  \centering
  \includegraphics[width=0.8\linewidth]{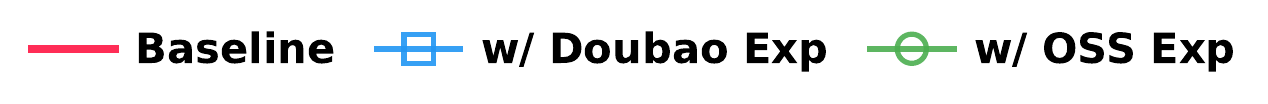}
  \begin{minipage}[t]{0.49\linewidth}
    \centering
    \includegraphics[width=\linewidth]{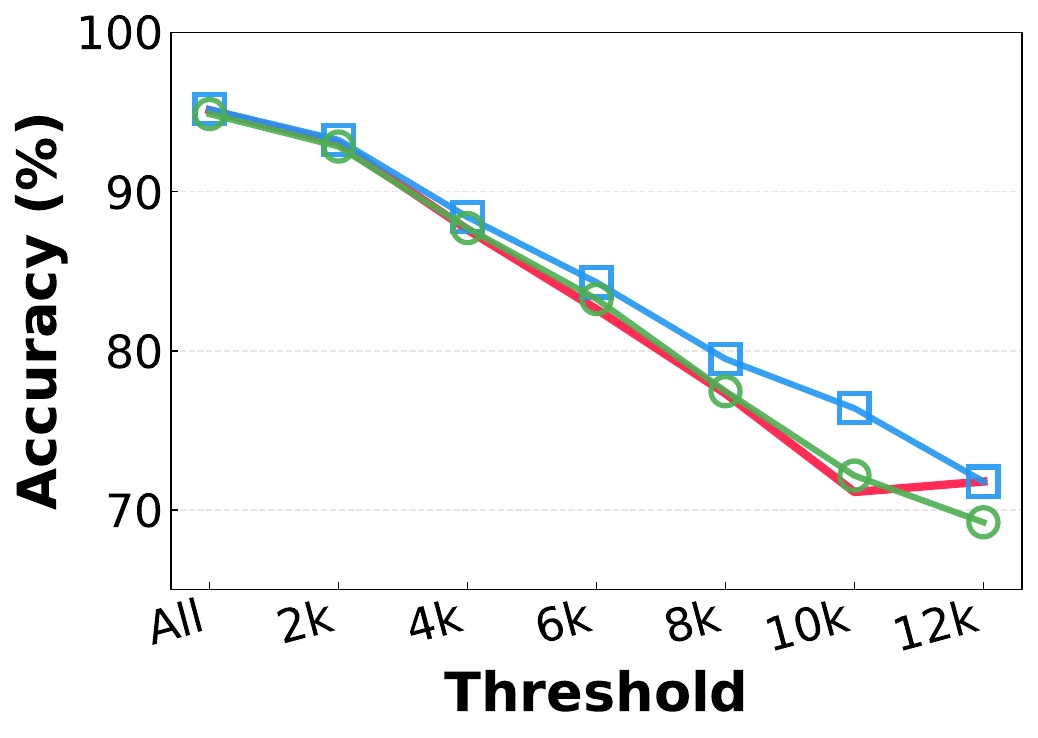}
    \subcaption{Doubao: Accuracy (\%)}
  \end{minipage}\hfill
  \begin{minipage}[t]{0.49\linewidth}
    \centering
    \includegraphics[width=\linewidth]{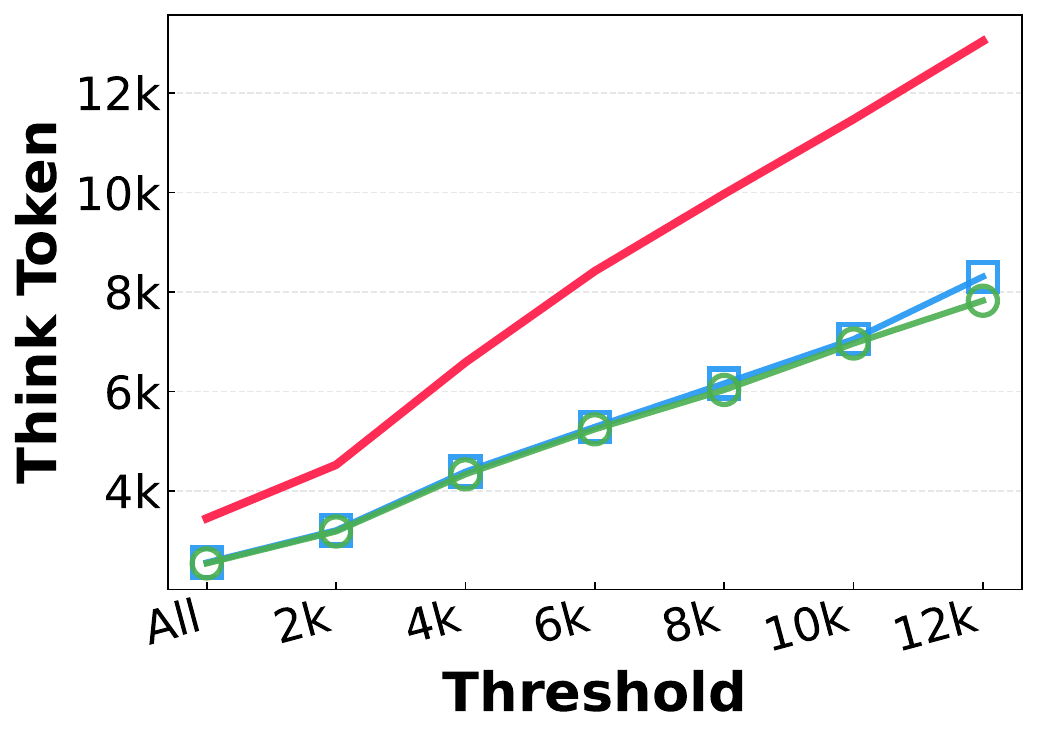}
    \subcaption{Doubao: Think Tokens}
  \end{minipage}\hfill

    \includegraphics[width=0.9\linewidth]{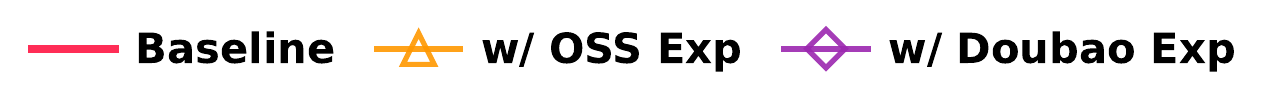}
  \begin{minipage}[t]{0.49\linewidth}
    \centering
    \includegraphics[width=\linewidth]{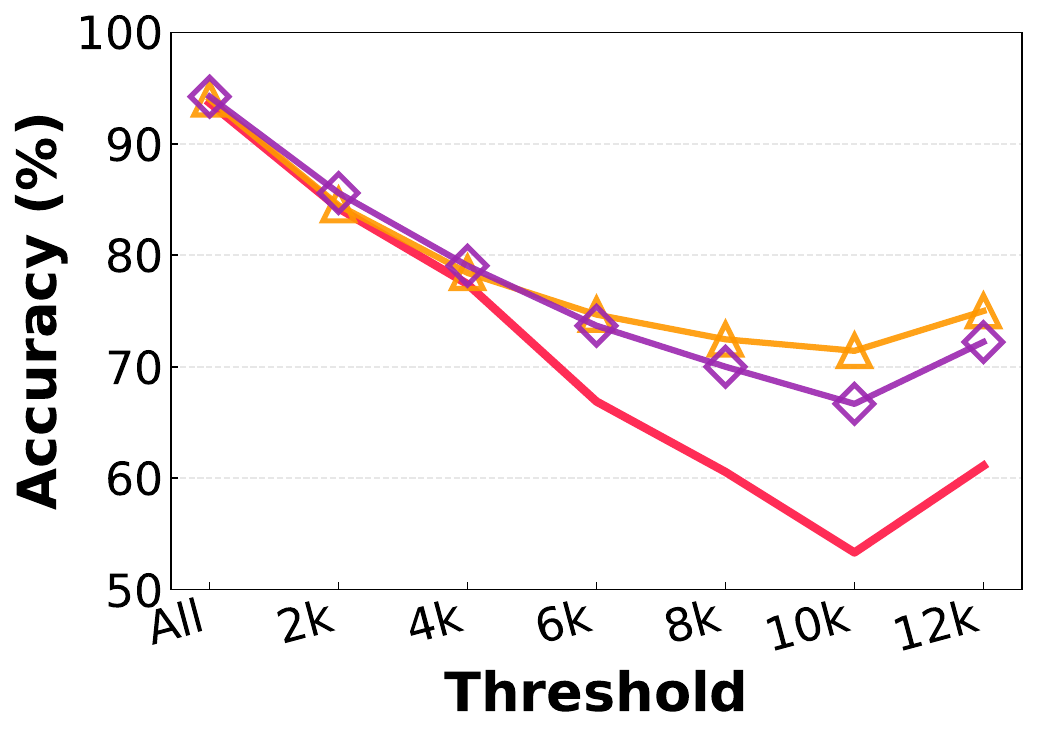}
    \subcaption{OSS: Accuracy (\%)}
  \end{minipage}\hfill
  \begin{minipage}[t]{0.49\linewidth}
    \centering
    \includegraphics[width=\linewidth]{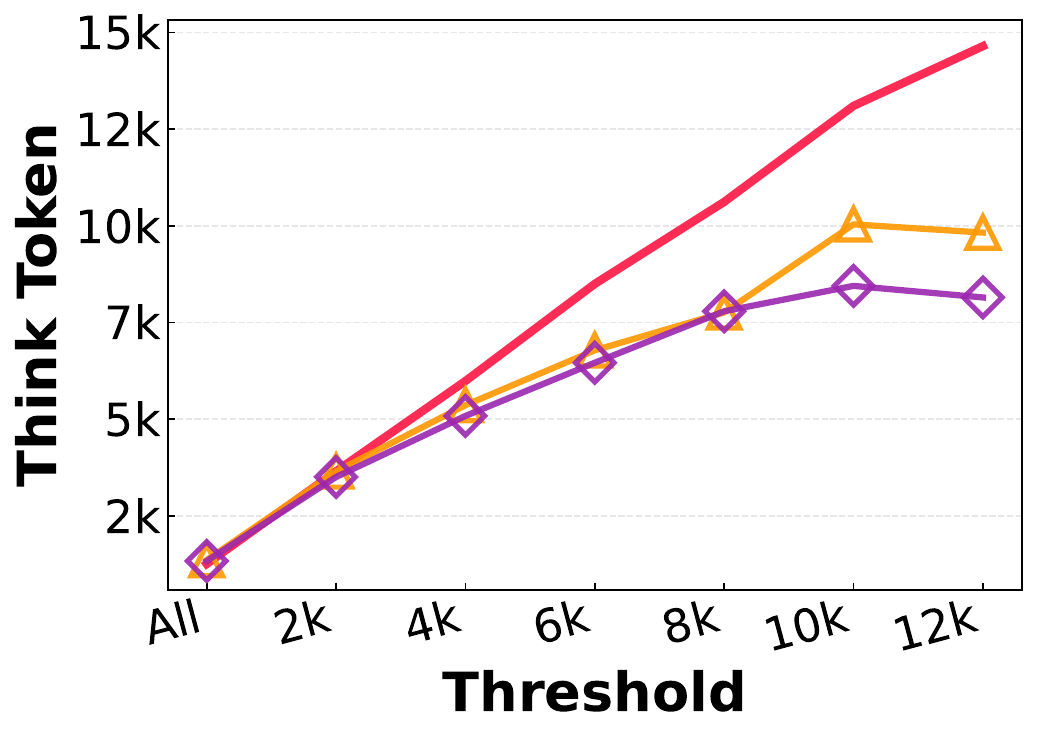}
    \subcaption{OSS: Think Tokens}
  \end{minipage}
  \caption{Cross-model transfer: Skills from Doubao vs. OSS applied to target models.}
  \label{fig:cross-comparison-6}
\end{figure}

We investigate whether skills distilled from one model (Source) can transfer effectively to another (Target) using Doubao- and OSS-based libraries. Results in Figure~\ref{fig:cross-comparison-6} show:

\begin{itemize}
    \item \textbf{Impact on Hard Problems:} The gains from TRS grow with problem difficulty $\theta$. As baseline accuracy declines, TRS better preserves performance while further reducing cost, suggesting that retrieved skills effectively prune redundant exploration.
    \item \textbf{Source Alignment:} Skills are highly transferable across models. However, \emph{same-source} skills (e.g., a Doubao library used for Doubao) generally yield the highest accuracy, likely due to stylistic alignment. Interestingly, \emph{cross-source} skills can sometimes lead to more aggressive token reduction.
    \item \textbf{Implication:} These results support a modular workflow: skills distilled from strong proprietary models can be used to improve more efficient deployment models, while model-specific distillation remains preferable when maximizing accuracy is the priority.
\end{itemize}

\subsection{Prompt Strategy for Skill Injection}
\label{sec:ablation-prompt}
\begin{figure}[tbp]
  \centering
  \includegraphics[width=\linewidth]{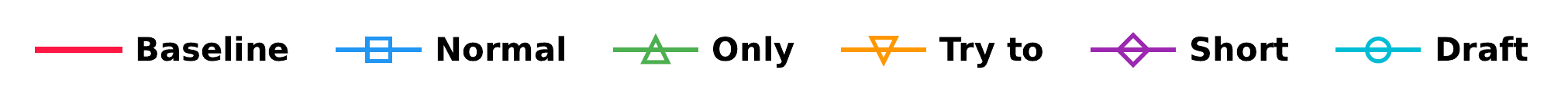}
  \vspace{0.1em}
  \begin{minipage}[t]{0.49\linewidth}
    \centering
    \includegraphics[width=\linewidth]{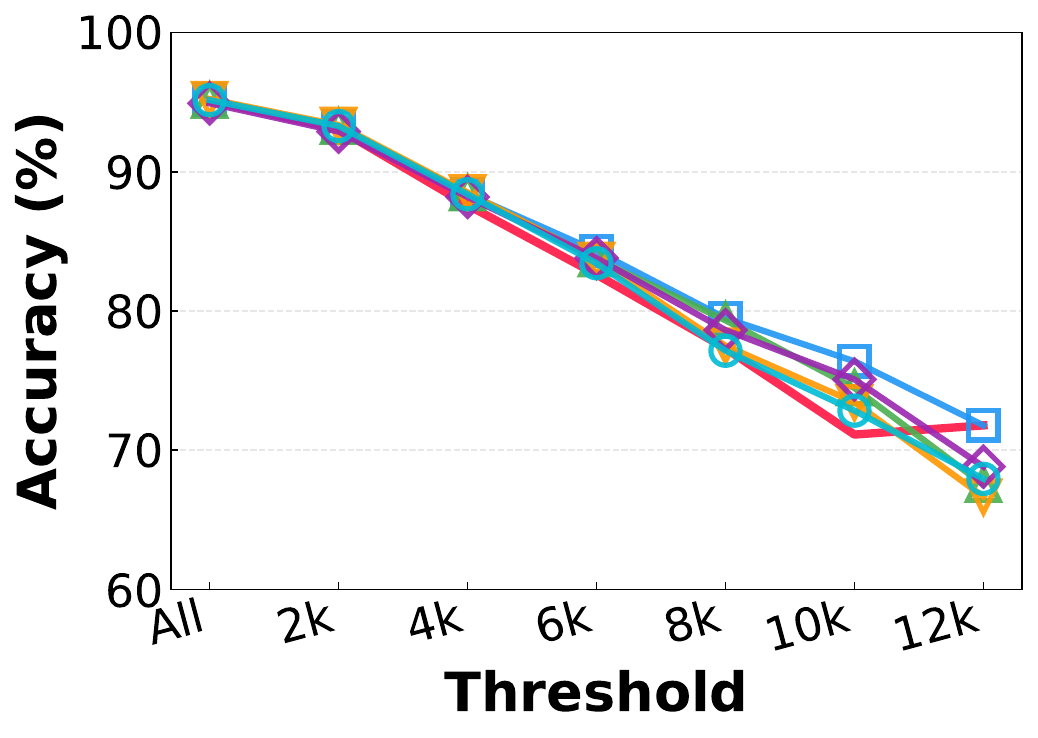}
    \subcaption{Doubao: Accuracy (\%)}
  \end{minipage}\hfill
  \begin{minipage}[t]{0.49\linewidth}
    \centering
    \includegraphics[width=\linewidth]{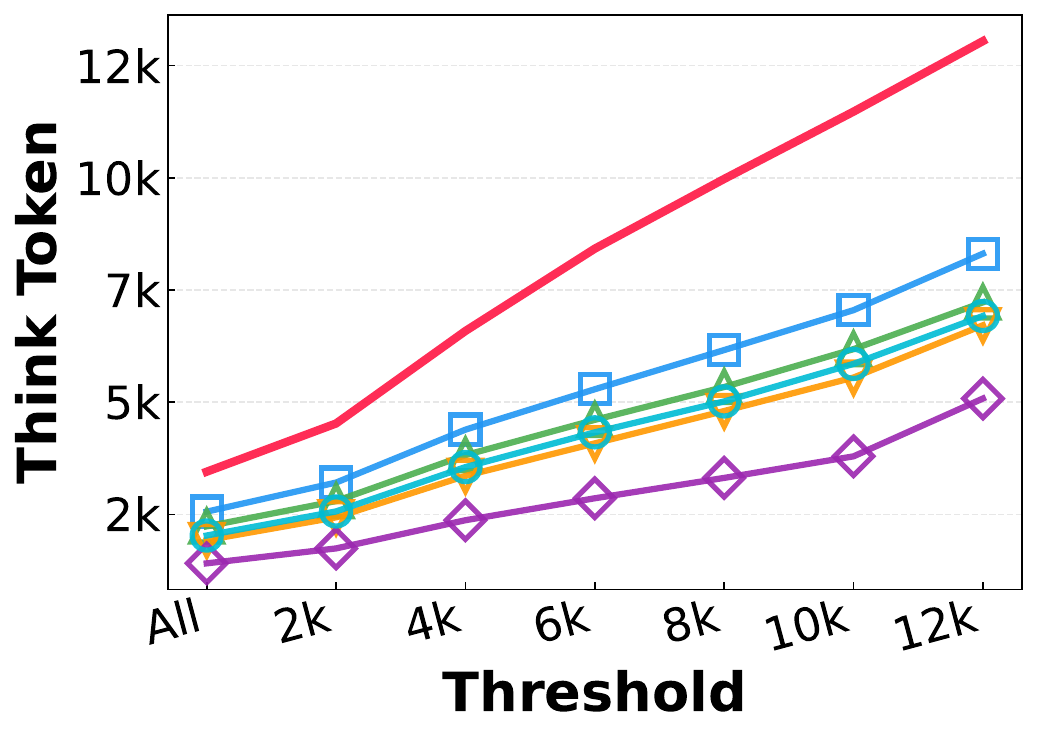}
    \subcaption{Doubao: Think Tokens}
  \end{minipage}

  \begin{minipage}[t]{0.49\linewidth}
    \centering
    \includegraphics[width=\linewidth]{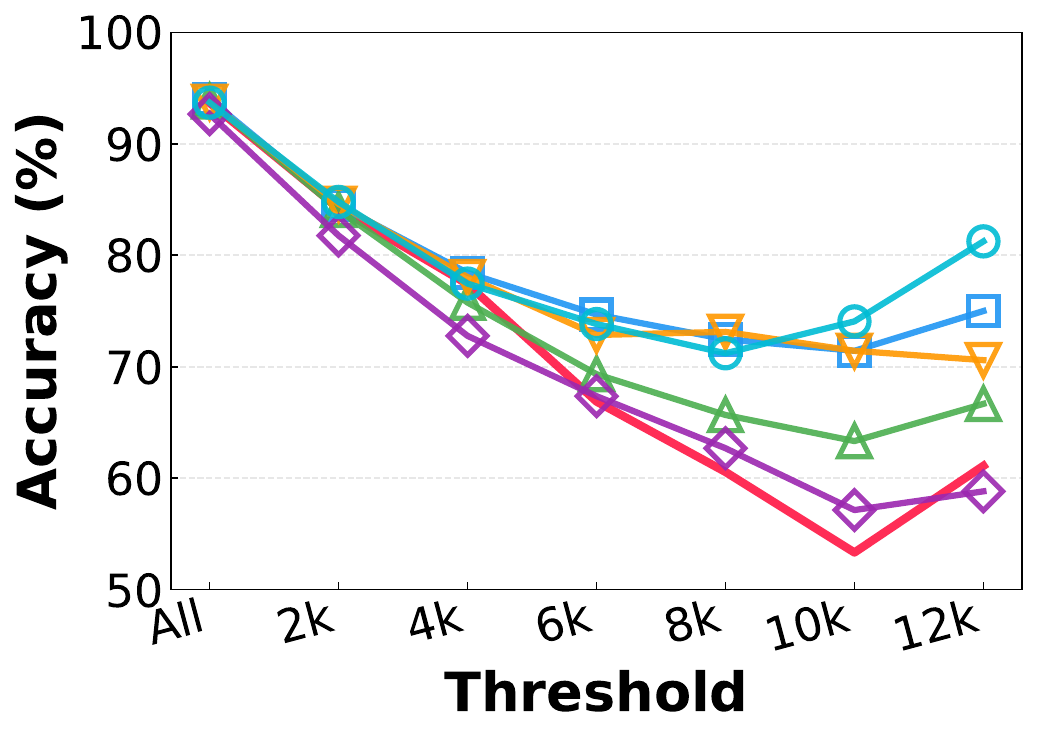}
    \subcaption{OSS: Accuracy (\%)}
  \end{minipage}\hfill
  \begin{minipage}[t]{0.49\linewidth}
    \centering
    \includegraphics[width=\linewidth]{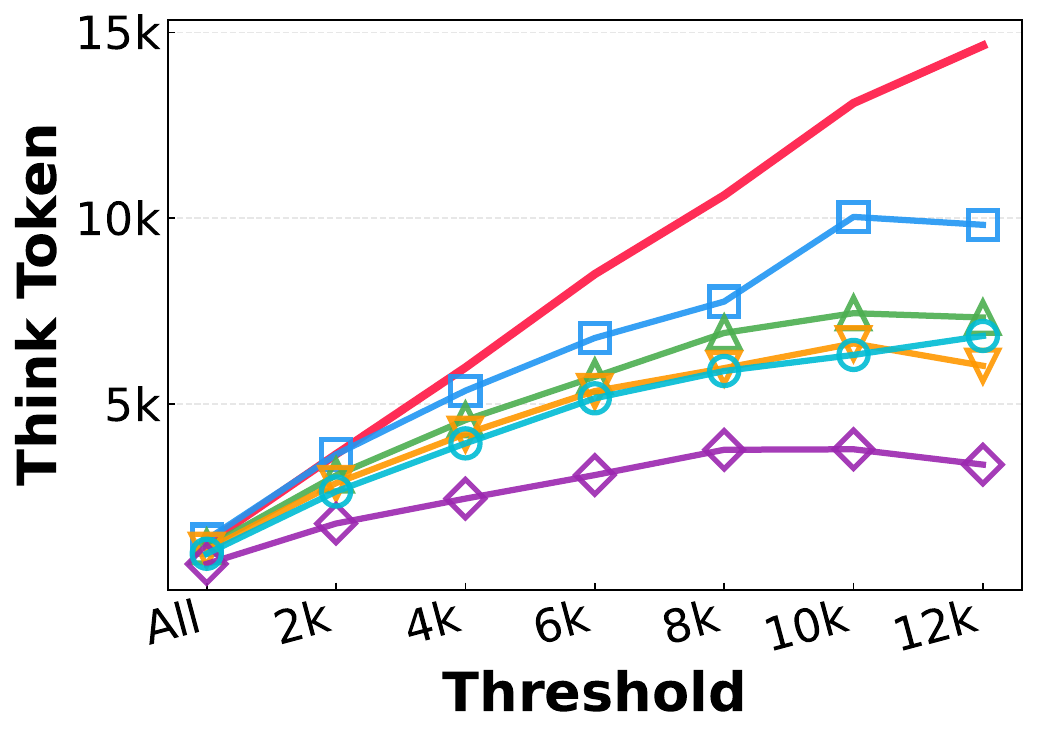}
    \subcaption{OSS: Think Tokens}
  \end{minipage}
  \caption{Impact of skill-injection prompts on accuracy and efficiency.}
  \label{fig:prompt-strategy-6}
\end{figure}

We ablate five prompt strategies (Normal, Only, Try to, Short, and Draft) to study the trade-off between accuracy and response length (Figure~\ref{fig:prompt-strategy-6}).

\begin{itemize}
    \item \textbf{Doubao:} The \textbf{Short} prompt, which imposes an explicit budget, provides the best trade-off, achieving substantial token reduction while maintaining competitive accuracy. We therefore adopt it as the default for Doubao.
    \item \textbf{OSS:} The \textbf{Draft} prompt, which encourages a concise step-by-step style, achieves the highest accuracy while retaining strong efficiency. In contrast, the \textbf{Short} prompt leads to noticeably larger accuracy degradation. We therefore use \textbf{Draft} as the default for OSS.
\end{itemize}
Exact templates are provided in Appendix~\ref{app:prompt-templates}.

\section{Summarizer-Choice Ablation}
\label{sec:summarizer-choice}

As an additional check beyond the representation controls in Table~\ref{tab:repr-coverage}, we regenerated the free-summary library with GPT-5-mini on the same fixed 500-question DeepMath subset. This variant reaches $89.4\%$ on Doubao and $89.6\%$ on OSS. Compared with the original free-summary row in Table~\ref{tab:repr-coverage}, the alternative summarizer improves OSS but weakens Doubao, and it still does not match full TRS on OSS. The result suggests that summarizer choice matters, but the final TRS effect is not reducible to a single free-form summary backend.

\section{Conclusion}

In this paper, we introduced \textbf{Thinking with Reasoning Skills (TRS)}, a framework that addresses the high inference cost of reasoning models by shifting the paradigm from ``reasoning from scratch'' to ``reasoning with recalled experience.'' 
Instead of forcing models to be brief—which often harms performance on complex tasks—TRS retrieves distilled, actionable \emph{skill cards} (encompassing both successful strategies and failure-mode fixes) to guide the model through the solution space. 
Empirical results on large-scale mathematical and coding benchmarks demonstrate that TRS successfully breaks the efficiency--accuracy trade-off: it significantly reduces reasoning tokens and overall inference costs while matching or exceeding the accuracy of standard Chain-of-Thought baselines, particularly on harder problems.
As a training-free, black-box compatible approach that generalizes across models, TRS offers a practical and scalable pathway for deploying capability-intensive reasoning models in real-world applications.

\section*{Limitations}
First, our analysis of \emph{failure cases} is not yet as deep as our success-skill study.
While TRS can distill ``failure-mode fixes'' from incorrect trajectories, our current pipeline relies on limited supervision signals (e.g., final answer correctness for math, and pass/fail information from test cases for coding).
We do not yet fully characterize which types of errors are most recoverable via skill distillation, nor do we have a principled method to summarize fine-grained failure dynamics beyond these signals.

Second, TRS introduces an additional retrieval and prompt-injection layer.
Although it is lightweight compared to long test-time reasoning, the end-to-end benefit depends on (i) retrieval quality and (ii) controlling prompt inflation.
In some settings (e.g., when completion tokens do not decrease or prompts grow too long), cost may not strictly improve even if accuracy increases.

Third, our current evaluation is still bounded. We focus mainly on math and competitive programming with automatic verifiers, plus targeted transfer and stress tests. Broader domains, weakly verifiable tasks, and open-ended workflows remain important future settings.

Finally, TRS opens a large design space. Future work can study which model should generate the original CoT, how to choose the summarizer, how to design the summarization and skill-use prompts, how to extend skill reuse to more domains, how to compress many problem-level skills into fewer general skills, and how to retrieve the most suitable skill from a very large library. The last question is especially connected to retrieval-augmented generation: as the skill library grows, retrieval design will become central to making experience reuse reliable and scalable.

\section*{Ethical Considerations}
This work adheres to ACL's ethical guidelines, and we state that there are no ethical concerns to our knowledge.

\section*{Acknowledgements}
We thank the anonymous reviewers, ACs, SACs, and PCs for their contributions.


\bibliography{custom}

\newpage

\appendix

\section{Skill Card Schema, Validation, and Distillation}
\label{app:skillformat}

\paragraph{Schema.}
Each skill is a compact structured card with five fields:
\textbf{Trigger} (applicability cues),
\textbf{Do} (minimal actionable recipe),
\textbf{Avoid} (anti-patterns),
\textbf{Check} (must-verify constraints/invariants),
\textbf{Risk} (edge cases/failure modes).
We use the same schema for math/coding and for both successes and failures.

\paragraph{Distillation input and rule.}
For each experience record $(q,a,\tau,y,c)$, the summarizer produces a card $s$ and 10--20 retrieval triggers $K$.
If $c{=}1$, $s$ summarizes the essential solution pattern.
If $c{=}0$, $s$ summarizes an anti-pattern and a concrete fix that generalizes beyond the instance.

\paragraph{Abstraction and safety constraints.}
We enforce three constraints:
(i) \textbf{abstraction}: avoid copying instance-specific constants, answers, or full code;
(ii) \textbf{actionability}: procedures must be executable as steps/checks;
(iii) \textbf{compactness}: each field is short (few bullets).
Cards violating schema or copying instance-specific details are discarded.
We also discard cards with missing fields or malformed structure.
The released code package includes the distillation, direct-inference, TRS-inference, and coding-specific prompt templates used to reproduce these steps.

\section{Retrieval Backends and Hybrid Fusion}
\label{app:retrieval}

\paragraph{Index keys.}
We index each card with $x=\mathrm{Concat}(q,K)$, where $K$ are summarizer-generated triggers.

\paragraph{Sparse (BM25).}
BM25 over $\{x\}$ provides strong lexical matching when problems share surface triggers (common in math).

\paragraph{Dense (BGE-M3 + FAISS).}
We embed queries and keys with BGE-M3 and run FAISS nearest-neighbor search.
Dense retrieval improves semantic matching when lexical overlap is weak (common in coding).

\paragraph{Hybrid retrieval.}
For coding, we use dual-route retrieval: BM25 candidates + dense candidates, merged with a simple score fusion and then truncated to top-$k$.
We ablate BM25-only / dense-only / hybrid and $k$ in Section~\ref{sec:ablation}.
Default choices: math uses BM25 ($k{=}1$); coding uses hybrid ($k{=}5$).

\begin{table*}[htbp]
\centering
\small
\caption{Comparison of retrieval strategies across mathematical reasoning and competitive programming. Math: 1,000 problems, Doubao-Seed-2.0 as generator, GPT-5-mini as verifier. Code: 1,000 holdout problems, Gemini-3-Flash as the inference model. ``$\Delta$'' denotes the difference against the no-retrieval baseline.}
\label{tab:retrieval-strategies}
\begin{tabular}{llccc}
\toprule
Domain & Retrieval & Accuracy (\%) & $\Delta$ (\%) & Avg Output Tok. \\
\midrule
\multirow{4}{*}{Math}
& None (Baseline) & 95.9 & --   & 2,240 \\
& BM25            & 95.5 & $-0.4$ & 2,240 \\
& Embedding       & 94.8 & $-1.1$ & 2,268 \\
& Hybrid          & 95.2 & $-0.7$ & 2,311 \\
\midrule
\multirow{3}{*}{Code}
& None (Baseline) & 72.0 & --    & 20,206 \\
& BM25            & 71.5 & $-0.5$ & 16,633 \\
& Hybrid          & 72.0 & $\pm 0.0$ & 16,491 \\
\bottomrule
\end{tabular}
\end{table*}

\section{Coding Pipeline: Data Filtering, Partitions, and Card Statistics}
\label{app:coding}

\paragraph{Dataset and split.}
We use \textsc{Nemotron-Competitive-Programming-v1} (34K).
We filter to 26,641 training instances with reliable local judging signals and hold out 1,000 additional instances for evaluation.
The 1,000-problem holdout is stratified into both\_ac (633), flash\_ac\_doubao\_wrong (95), flash\_wrong\_doubao\_ac (21), and both\_wrong (251).

\paragraph{Step 1: Data preparation and filtering.}
We retain instances that are consistently judged by our local compile-and-run evaluator and exclude instances with unstable verdicts (e.g., ambiguous I/O, inconsistent tests, or frequent runtime issues).
This filtering reduces noise in both card distillation and final evaluation.

\paragraph{Step 2: Outcome partitioning (four types).}
We run two generators (Gemini 3 Flash Preview; Doubao Seed 2.0 Pro) and categorize each training instance by verdict:
both\_ac (16,851),
flash\_ac\_doubao\_wrong (2,521),
flash\_wrong\_doubao\_ac (570),
both\_wrong (6,699).
These categories produce success experience, contrastive paired experience, and failure experience.

\paragraph{Step 3: Experience distillation (cards).}
We distill structured cards from trajectories using Gemini Flash.
The v5 coding library released with the camera-ready artifacts contains 42,564 compact cards:
success (31,044), contrast (5,069), and failure diagnostics (6,451), where the failure subset consists of wrong-approach cards (3,733) and edge-fix cards (2,718).
To reduce redundancy and prompt inflation at inference time, we keep success cards from only one generator model, while preserving contrast/failure cards.

\paragraph{Step 4: Index construction.}
We build a dual-route retrieval index:
BM25 over text keys and FAISS over dense embeddings (BGE-M3).

\paragraph{Step 5: Inference injection and evaluation.}
On the 1K test split, we retrieve top-5 cards and inject them into the prompt.
The model receives only (retrieved cards + problem statement), with no additional reasoning-style prompting.
Correctness is determined by local compile-and-run on test cases (pass@1).

\section{Local Coding Judge and Execution Environment}
\label{app:judge}

\paragraph{Judge protocol.}
For each generated program, we compile and run it against the provided test cases and compare outputs.
A solution is correct iff it passes all tests (pass@1).

\paragraph{Environment.}
We use a fixed, sandboxed environment for all methods to ensure fairness.
The released judge extracts Python/C++ code, compiles C++ with \texttt{g++ -O2 -std=gnu++17 -pipe}, runs Python with the local Python interpreter, applies a default 2-second timeout and 1024MB memory cap unless a problem specifies its own limits, and terminates non-terminating processes.
The public release should be run inside an isolated sandbox because executing generated programs is inherently unsafe.

\section{BM25 vs Embedding}
Table \ref{tab:retrieval-strategies} shows the result.

\paragraph{Mathematical Reasoning.}
On the math benchmark, all three retrieval methods cause a slight accuracy drop compared to the no-retrieval baseline (95.9\%).
BM25 incurs the smallest degradation ($-0.4$\%), whereas dense embedding retrieval suffers the most ($-1.1$\%) and hybrid falls in between ($-0.7$\%).
We attribute this to the fact that mathematical heuristics are best matched via lexical cues:
BM25 excels at surface-level matching of formula patterns and theorem names, while embedding models---trained predominantly on natural language---tend to retrieve semantically plausible but structurally mismatched hints that occasionally mislead the generator.
Meanwhile, token overhead is nearly identical across methods ($\sim$2{,}240--2{,}311 output tokens per problem).
Given its best accuracy and lowest cost, we adopt pure BM25 for math retrieval.

\paragraph{Competitive Programming.}
On the coding benchmark, the pattern reverses: hybrid retrieval fully preserves baseline accuracy (72.0\%), while BM25 alone shows a minor drop ($-0.5$\%).
This is expected because algorithmic problems benefit from semantic similarity---problems with different surface descriptions may share underlying algorithmic patterns that dense embeddings capture but keyword matching misses.
Both retrieval strategies substantially reduce output tokens ($\sim$16{,}500 vs.\ 20{,}200), indicating that injected experience hints help the model generate more concise solutions.

\section{Token and Cost Accounting}
\label{app:cost}

\paragraph{Tokens.}
When explicit think tokens are exposed, we report thinking length directly.
Otherwise, we report output token length as a proxy, applied consistently across compared methods.

\paragraph{Cost proxy.}
We compute cost from input and output token counts under a fixed input/output token price ratio for each model, but report only normalized percentages in the paper. Direct is set to 100.0\% within each model--dataset pair, and TRS cost is reported relative to that baseline. A value below 100.0\% indicates lower estimated per-query cost; deltas such as $\downarrow 17.5\%$ denote percentage reduction against Direct. This normalization keeps cost comparisons readable while preserving the effect of TRS's longer prompt and shorter generated reasoning.

\section{Prompt Templates}
\label{app:prompt-templates}

We provide the exact prompt templates used in our experiments to facilitate reproducibility. These templates illustrate how reasoning skills are injected into the model context and how baseline constraints are enforced.

\begin{figure}[h!]
\setlength{\abovecaptionskip}{0.1cm}
\centering
\begin{tcolorbox}[
  title={TRS prompt: Normal},
  colback=gray!3, colframe=black!35,
  boxrule=0.6pt, arc=1.5mm,
  left=1.5mm,right=1.5mm,top=1mm,bottom=1mm
]
\small\ttfamily
You are a helpful and harmless assistant.

You may be given an optional Solving Hints section. Use it only if it is relevant to the problem; otherwise, ignore it completely.

[Solving Hints]
{SOLVING\_HINTS}
[/Solving Hints]

Problem:
{PROBLEM}
\end{tcolorbox}
\caption{Full prompt template for TRS-Normal.}
\label{fig:prompt-trs-normal}
\end{figure}

\begin{figure}[h!]
\setlength{\abovecaptionskip}{0.1cm}
\centering
\begin{tcolorbox}[title={TRS prompt: Only}, colback=gray!3, colframe=black!35, boxrule=0.6pt, arc=1.5mm,
  left=1.5mm,right=1.5mm,top=1mm,bottom=1mm]
\small\ttfamily
You are a helpful and harmless assistant.

You may be given an optional Solving Hints section. Use it only if it is relevant to the problem; otherwise, ignore it completely.

[Solving Hints]
{SOLVING\_HINTS}
[/Solving Hints]

\textcolor{red}{Only try to reduce the number of tokens used if the solution hints are useful; otherwise, please think normally.} Problem:
{PROBLEM}
\end{tcolorbox}
\caption{Full prompt template for TRS-Only.}
\label{fig:prompt-trs-only}
\end{figure}

\begin{figure}[h!]
\setlength{\abovecaptionskip}{0.1cm}
\centering
\begin{tcolorbox}[title={TRS prompt: Try-to}, colback=gray!3, colframe=black!35, boxrule=0.6pt, arc=1.5mm,
  left=1.5mm,right=1.5mm,top=1mm,bottom=1mm]
\small\ttfamily
You are a helpful and harmless assistant.

You may be given an optional Solving Hints section. Use it only if it is relevant to the problem; otherwise, ignore it completely.

[Solving Hints]
{SOLVING\_HINTS}
[/Solving Hints]

\textcolor{red}{If you use the solving hints, please try to reduce the number of tokens used.} Problem:
{PROBLEM}
\end{tcolorbox}
\caption{Full prompt template for TRS-Try-to.}
\label{fig:prompt-trs-tryto}
\end{figure}

\begin{figure}[h!]
\setlength{\abovecaptionskip}{0.1cm}
\centering
\begin{tcolorbox}[title={TRS prompt: Short (budgeted)}, colback=gray!3, colframe=black!35, boxrule=0.6pt, arc=1.5mm,
  left=1.5mm,right=1.5mm,top=1mm,bottom=1mm]
\small\ttfamily
You are a helpful and harmless assistant.

You may be given an optional Solving Hints section. Use it only if it is relevant to the problem; otherwise, ignore it completely.

[Solving Hints]
{SOLVING\_HINTS}
[/Solving Hints]

\textcolor{red}{Let's think step by step and use less than [budget] tokens:}
{PROBLEM}
\end{tcolorbox}
\caption{Full prompt template for TRS-Short (budgeted).}
\label{fig:prompt-trs-short}
\end{figure}

\begin{figure}[h!]
\setlength{\abovecaptionskip}{0.1cm}
\centering
\begin{tcolorbox}[title={TRS prompt: Draft (CoD-like)}, colback=gray!3, colframe=black!35, boxrule=0.6pt, arc=1.5mm,
  left=1.5mm,right=1.5mm,top=1mm,bottom=1mm]
\small\ttfamily
You are a helpful and harmless assistant.

You may be given an optional Solving Hints section. Use it only if it is relevant to the problem; otherwise, ignore it completely.

[Solving Hints]
{SOLVING\_HINTS}
[/Solving Hints]

\textcolor{red}{Think step by step, but only keep a minimum draft for each thinking step, with 5 words at most.} Problem:
{PROBLEM}
\end{tcolorbox}
\caption{Full prompt template for TRS-Draft (CoD-like).}
\label{fig:prompt-trs-draft}
\end{figure}

\subsection{TRS Skill-Injected Prompts}
\label{app:prompts-trs}

Figures~\ref{fig:prompt-trs-normal} through \ref{fig:prompt-trs-draft} present the different TRS skill-injection prompt variants evaluated in our ablation study (Section~\ref{sec:ablation-prompt}). These templates reflect different ways of incorporating retrieved skills into the model prompt, ranging from more permissive guidance to more explicit constraints on reasoning style and length.

We find that different models respond differently to such prompt constraints. Accordingly, based on the empirical results reported in Section~\ref{sec:ablation-prompt}, we adopt different default templates for different model families:
\begin{itemize}
    \item For the \textbf{Doubao} family, we adopt the \textbf{Short} template (Figure~\ref{fig:prompt-trs-short}) as the default, since it explicitly enforces a tighter budget and achieves the best efficiency--accuracy trade-off in our experiments.
    \item For \textbf{GPT-OSS-120B}, we select the \textbf{Draft} template (Figure~\ref{fig:prompt-trs-draft}) as the default, as it encourages a concise chain-of-draft style that better preserves accuracy while still reducing response length effectively.
\end{itemize}

\subsection{Baseline Prompts}
\label{app:prompts-baselines}

Figures~\ref{fig:prompt-tale-ep}, \ref{fig:prompt-cod}, and \ref{fig:prompt-no-wait} display the prompts used for the comparison methods. These baselines rely on prompt-engineering techniques to constrain reasoning length (TALE-EP, CoD) or suppress specific tokens (No-Wait) without the aid of external retrieved skills.

\begin{figure}[htbp]
\centering
\begin{tcolorbox}[title={TALE-EP prompt: Budget + Solve}, colback=gray!3, colframe=black!35, boxrule=0.6pt, arc=1.5mm,
  left=1.5mm,right=1.5mm,top=1mm,bottom=1mm]
\small\ttfamily
Question:\\
\{QUESTION\}

Let's think step by step and use less than [Budget Here] tokens:\\
----\\
Question:\\
\{QUESTION\}

Task: Analyze the given question and estimate the minimum number of tokens required to generate a complete and accurate response. Please give the response by strictly following this format: [[budget]], for example, Budget: [[12]].
\end{tcolorbox}
\caption{Full prompt template for TALE-EP (two-phase: budget estimation and solve).}
\label{fig:prompt-tale-ep}
\end{figure}

\begin{figure}[htbp]
\setlength{\abovecaptionskip}{-0cm}
\centering
\begin{tcolorbox}[title={CoD prompt}, colback=gray!3, colframe=black!35, boxrule=0.6pt, arc=1.5mm,
  left=1.5mm,right=1.5mm,top=1mm,bottom=1mm]
\small\ttfamily
Question:\\
\{QUESTION\}

Think step by step, but only keep a minimum draft for each thinking step, with 5 words at most. Return the answer at the end of the response after a separator \#\#\#\#.
\end{tcolorbox}
\caption{Full prompt template for CoD.}
\label{fig:prompt-cod}
\end{figure}

\begin{figure}[htbp]
\setlength{\abovecaptionskip}{-0cm}
\centering
\begin{tcolorbox}[title={No-Wait prompt}, colback=gray!3, colframe=black!35, boxrule=0.6pt, arc=1.5mm,
  left=1.5mm,right=1.5mm,top=1mm,bottom=1mm]
\small\ttfamily
Question:\\
\{QUESTION\}

Think step by step. Do not use any of the following words in your thinking process:\\
``wait'', ``alternatively'', ``hmm'', ``but'', ``however'', ``alternative'', ``another'', ``check'', ``double-check'', ``oh'', ``maybe'', ``verify'', ``other'', ``again'', ``now'', ``ah'', ``any''.
\end{tcolorbox}
\caption{Full prompt template for No-Wait.}
\label{fig:prompt-no-wait}
\end{figure}

\section{Extended Comparison with Chain-of-Draft (CoD)}
\label{app:more-results}

\begin{figure*}[h!]
\setlength{\abovecaptionskip}{0.1cm}
  \centering
  \includegraphics[width=0.4\linewidth]{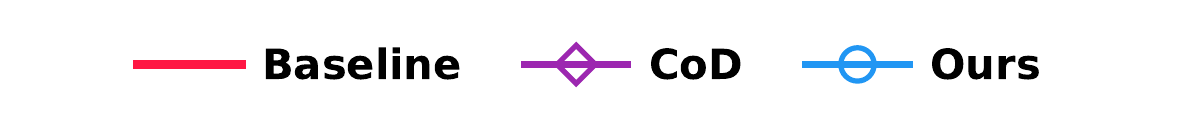}
  
  \begin{minipage}[t]{0.32\linewidth}
    \centering
    \includegraphics[width=\linewidth]{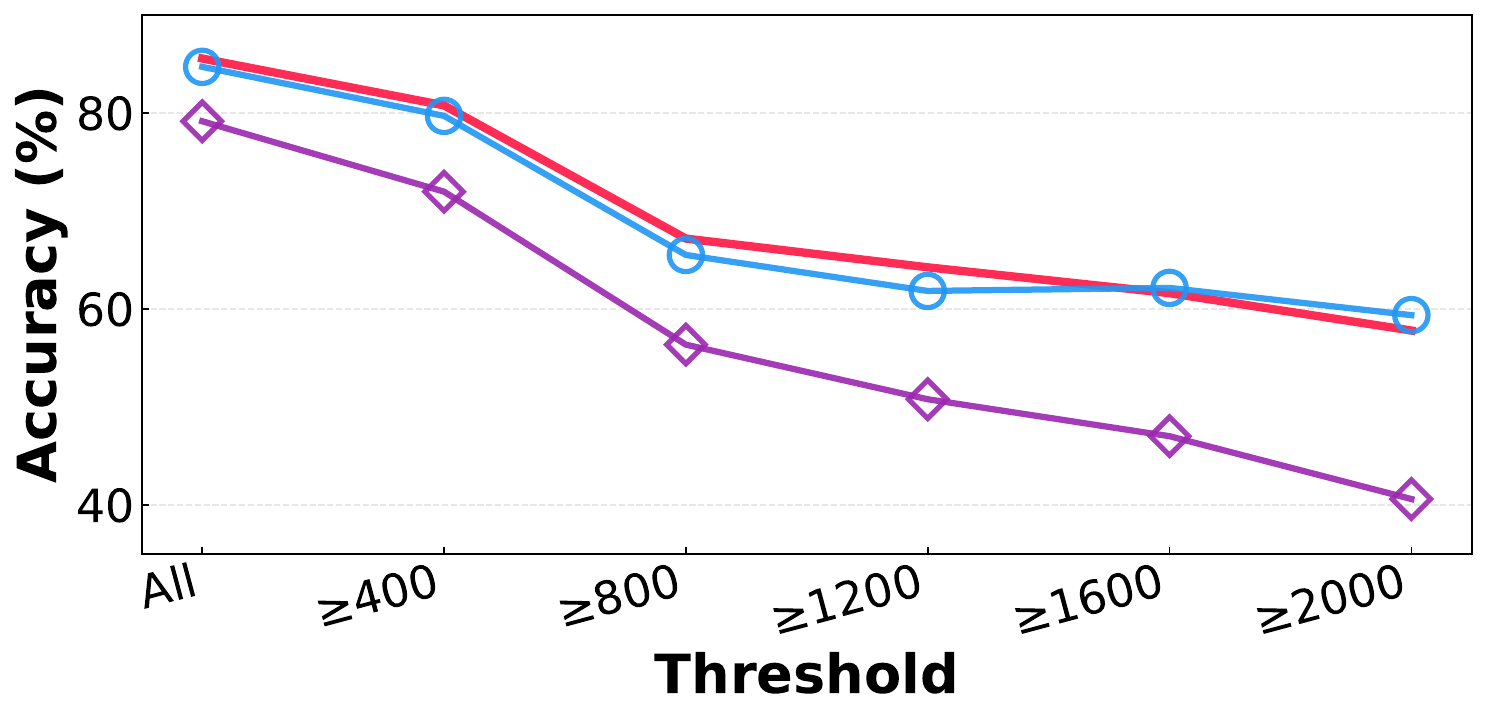}
    \subcaption{GPT-5.2: Acc.}
  \end{minipage}\hfill
  \begin{minipage}[t]{0.32\linewidth}
    \centering
    \includegraphics[width=\linewidth]{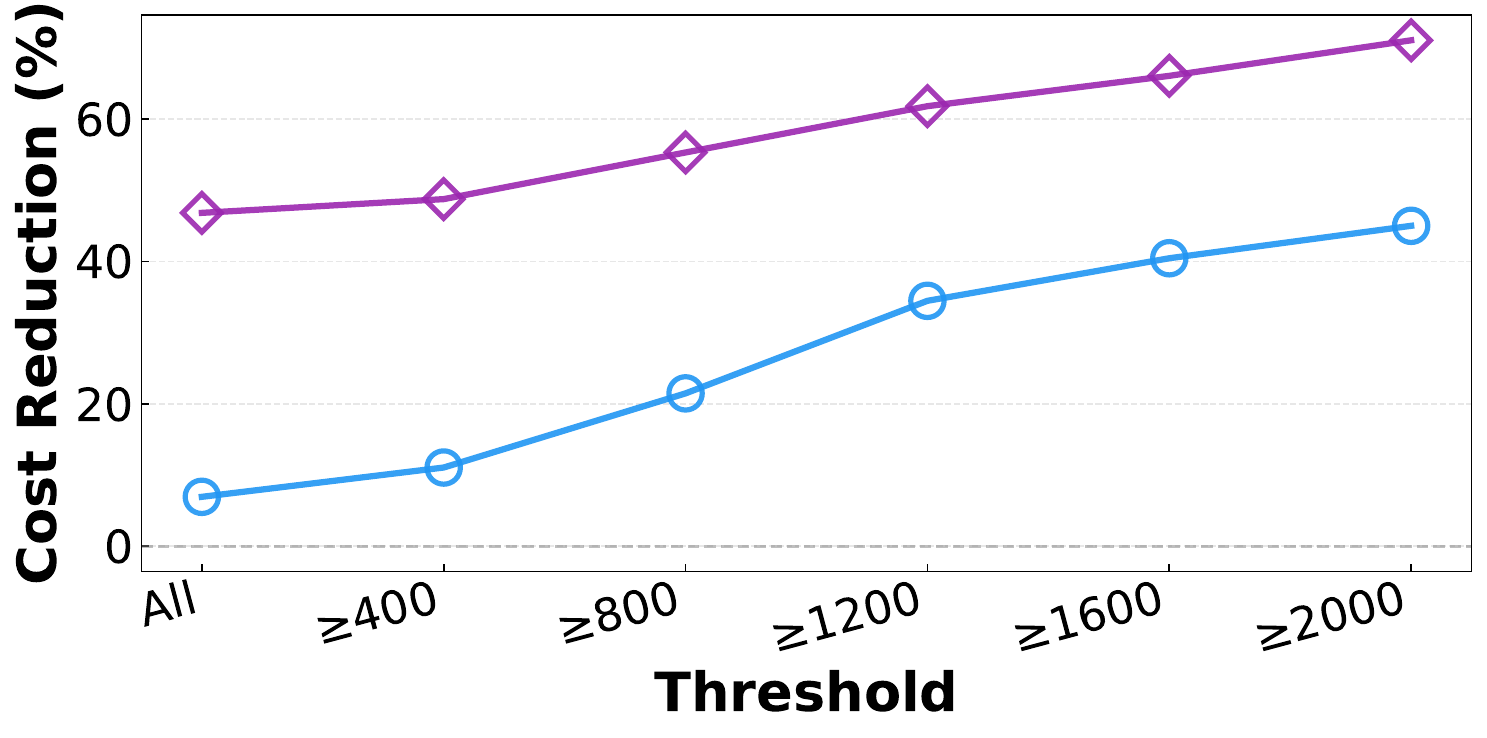}
    \subcaption{GPT-5.2: Cost}
  \end{minipage}\hfill
  \begin{minipage}[t]{0.32\linewidth}
    \centering
    \includegraphics[width=\linewidth]{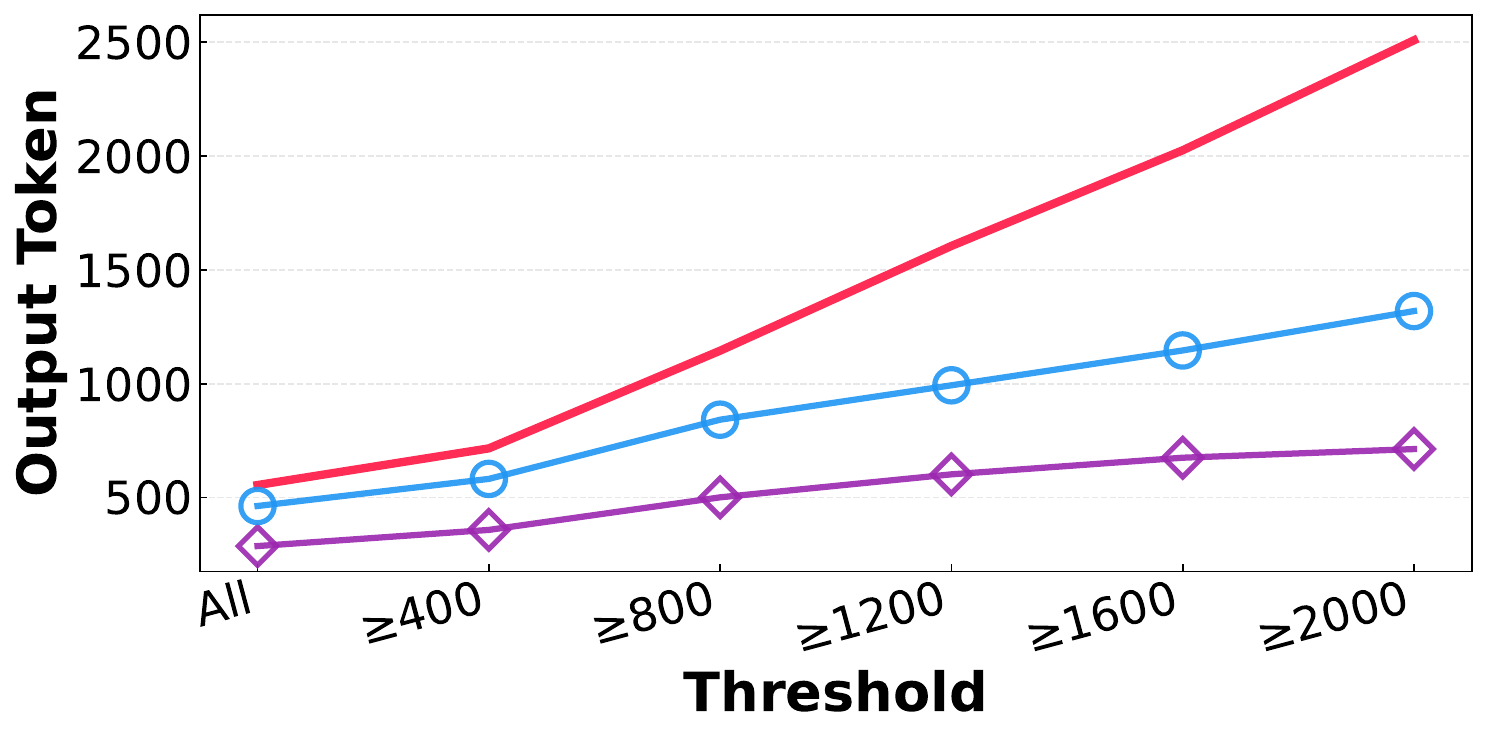}
    \subcaption{GPT-5.2: Tokens}
  \end{minipage}

  \vspace{0.5em}

  \begin{minipage}[t]{0.32\linewidth}
    \centering
    \includegraphics[width=\linewidth]{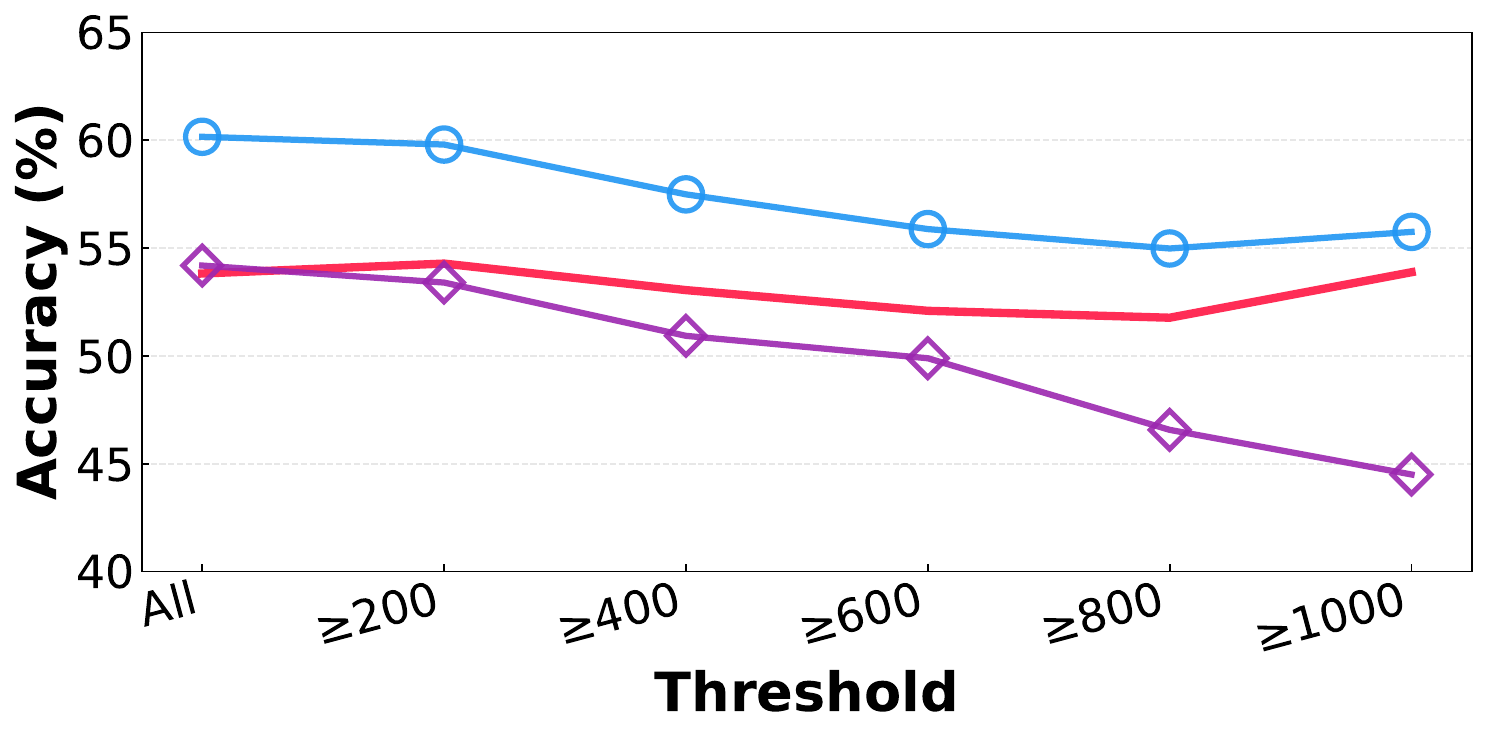}
    \subcaption{Grok-4 Fast: Acc.}
  \end{minipage}\hfill
  \begin{minipage}[t]{0.32\linewidth}
    \centering
    \includegraphics[width=\linewidth]{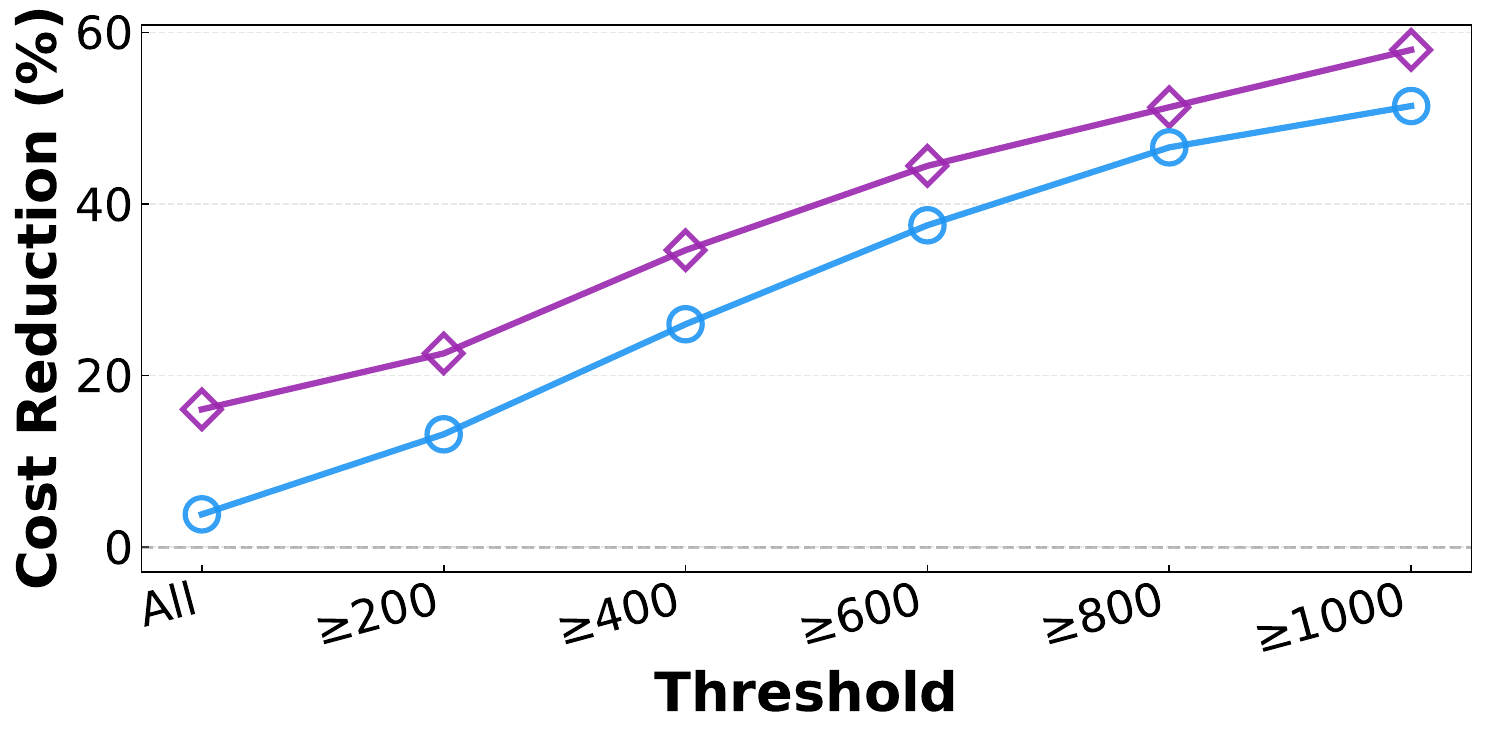}
    \subcaption{Grok-4 Fast: Cost}
  \end{minipage}\hfill
  \begin{minipage}[t]{0.32\linewidth}
    \centering
    \includegraphics[width=\linewidth]{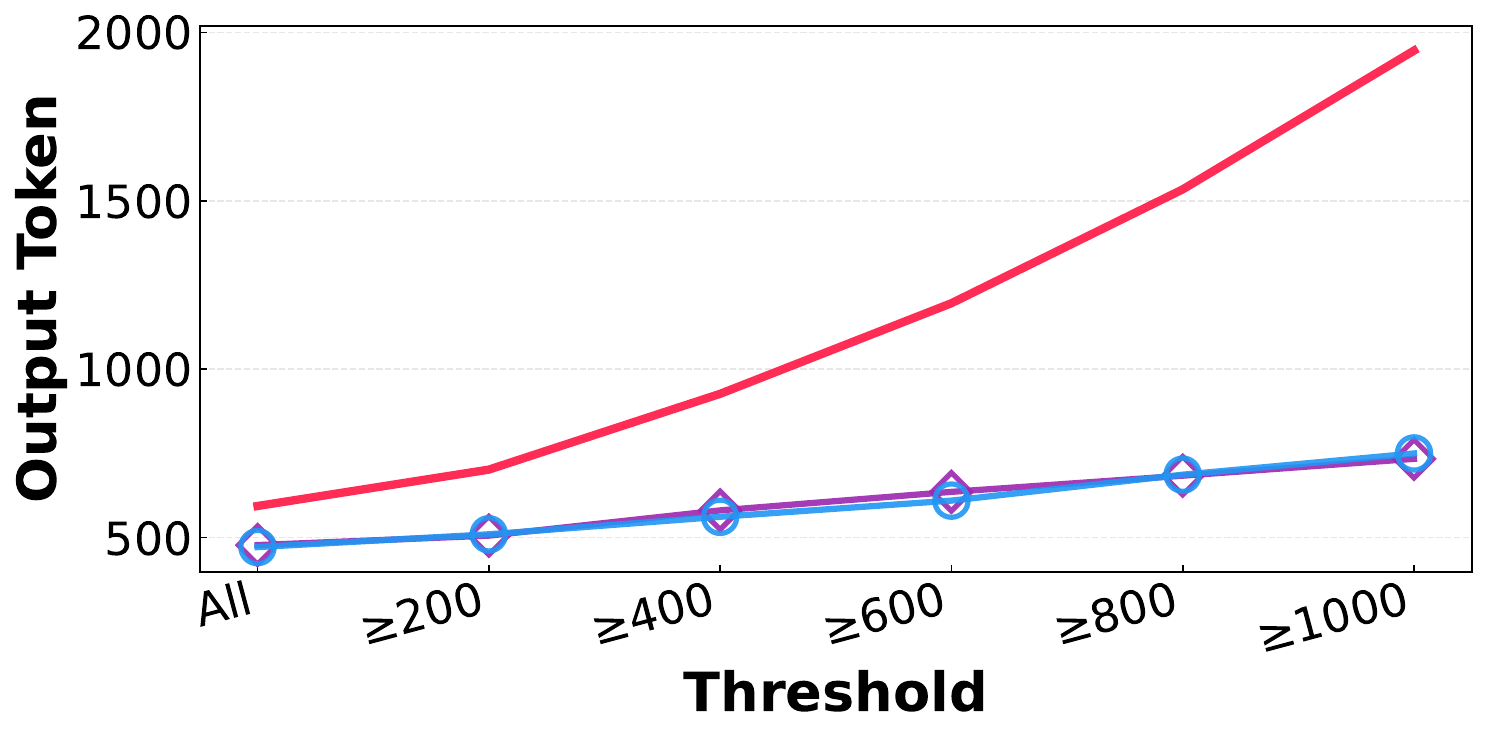}
    \subcaption{Grok-4 Fast: Tokens}
  \end{minipage}

  \vspace{0.5em}

  \begin{minipage}[t]{0.32\linewidth}
    \centering
    \includegraphics[width=\linewidth]{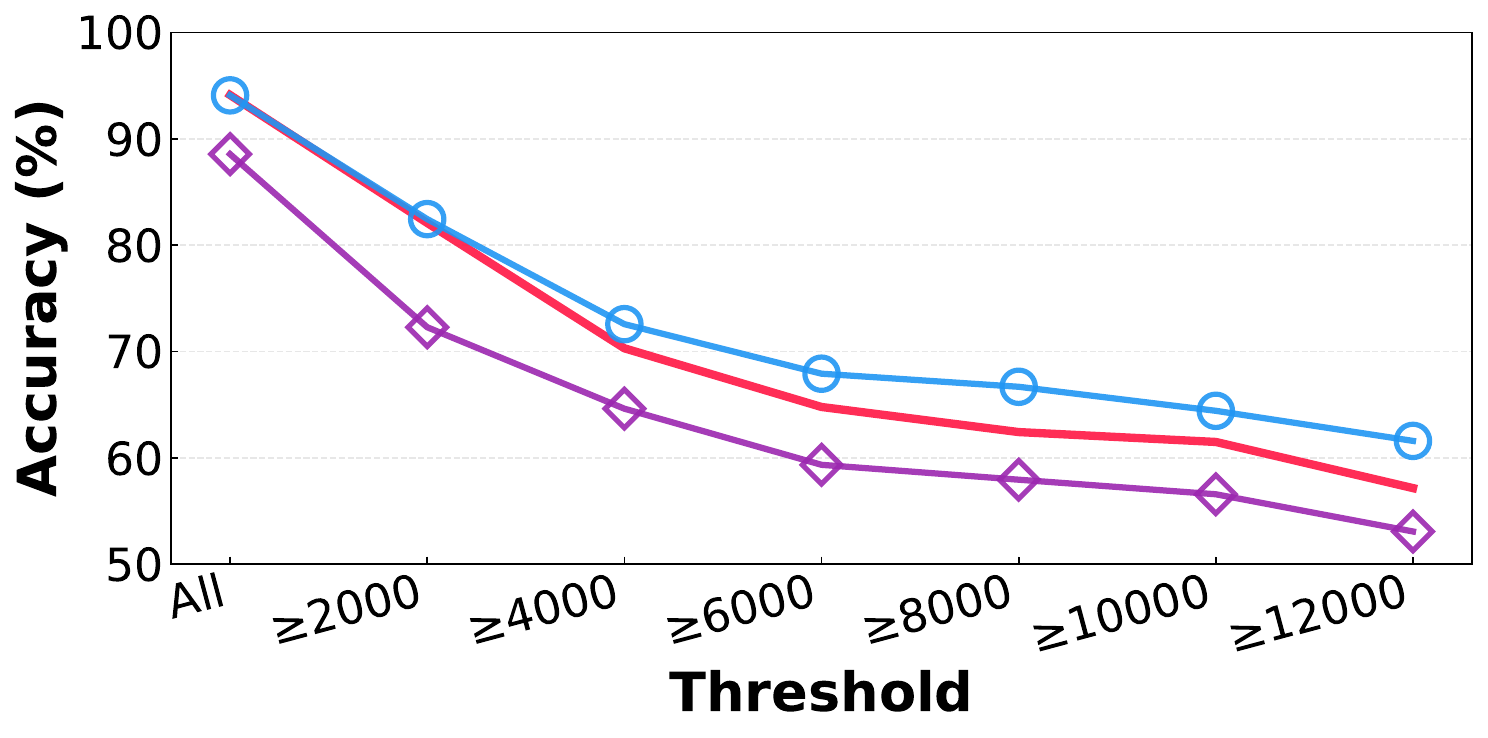}
    \subcaption{Gemini-3 Pro: Acc.}
  \end{minipage}\hfill
  \begin{minipage}[t]{0.32\linewidth}
    \centering
    \includegraphics[width=\linewidth]{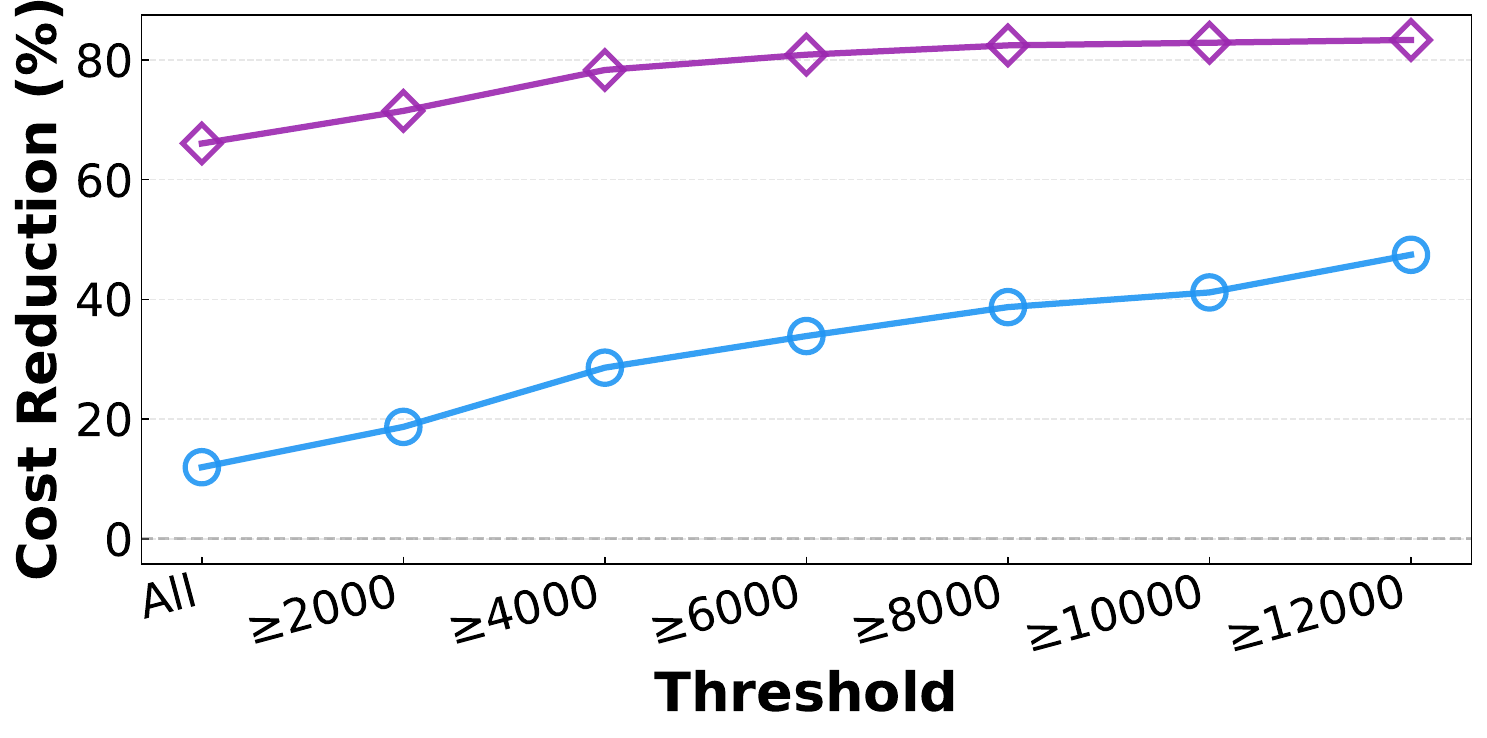}
    \subcaption{Gemini-3 Pro: Cost}
  \end{minipage}\hfill
  \begin{minipage}[t]{0.32\linewidth}
    \centering
    \includegraphics[width=\linewidth]{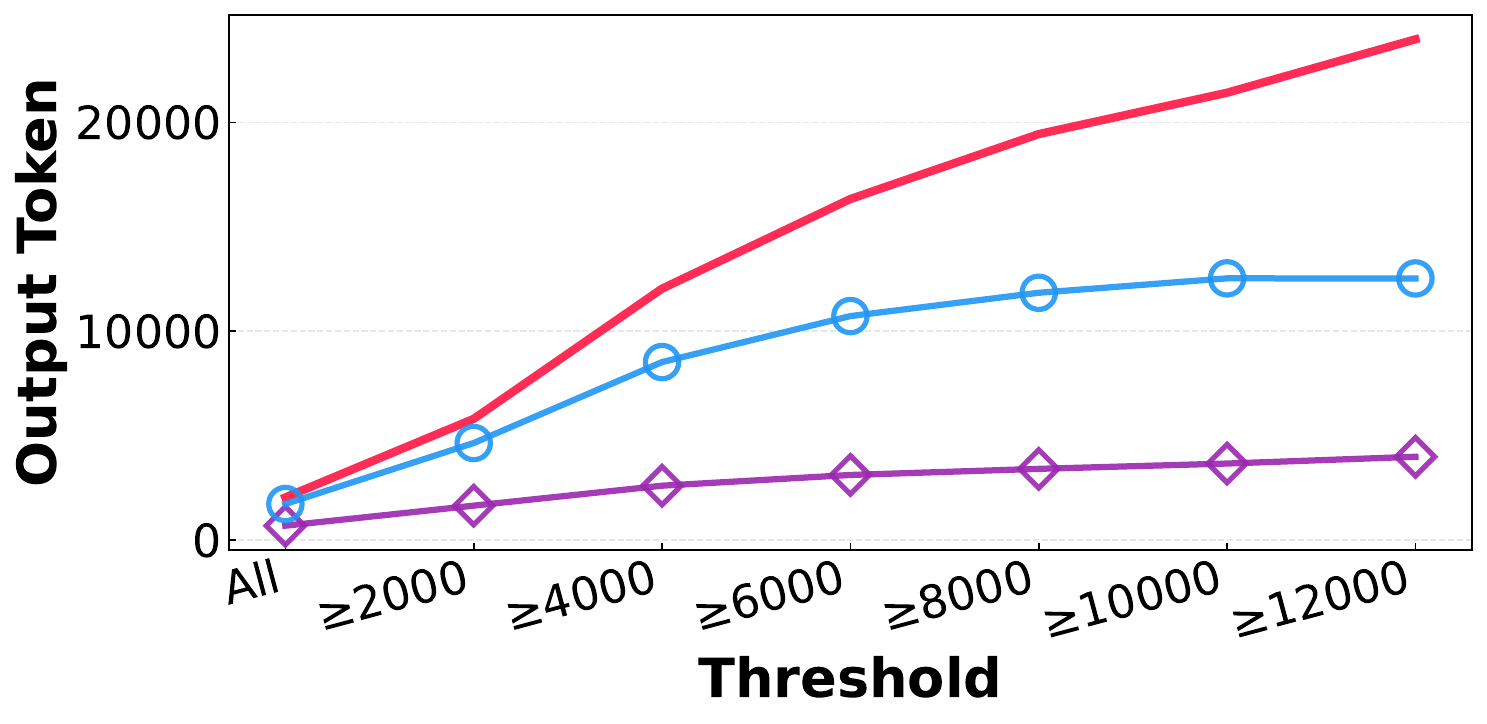}
    \subcaption{Gemini-3 Pro: Tokens}
  \end{minipage}

  \vspace{0.5em}

  \begin{minipage}[t]{0.32\linewidth}
    \centering
    \includegraphics[width=\linewidth]{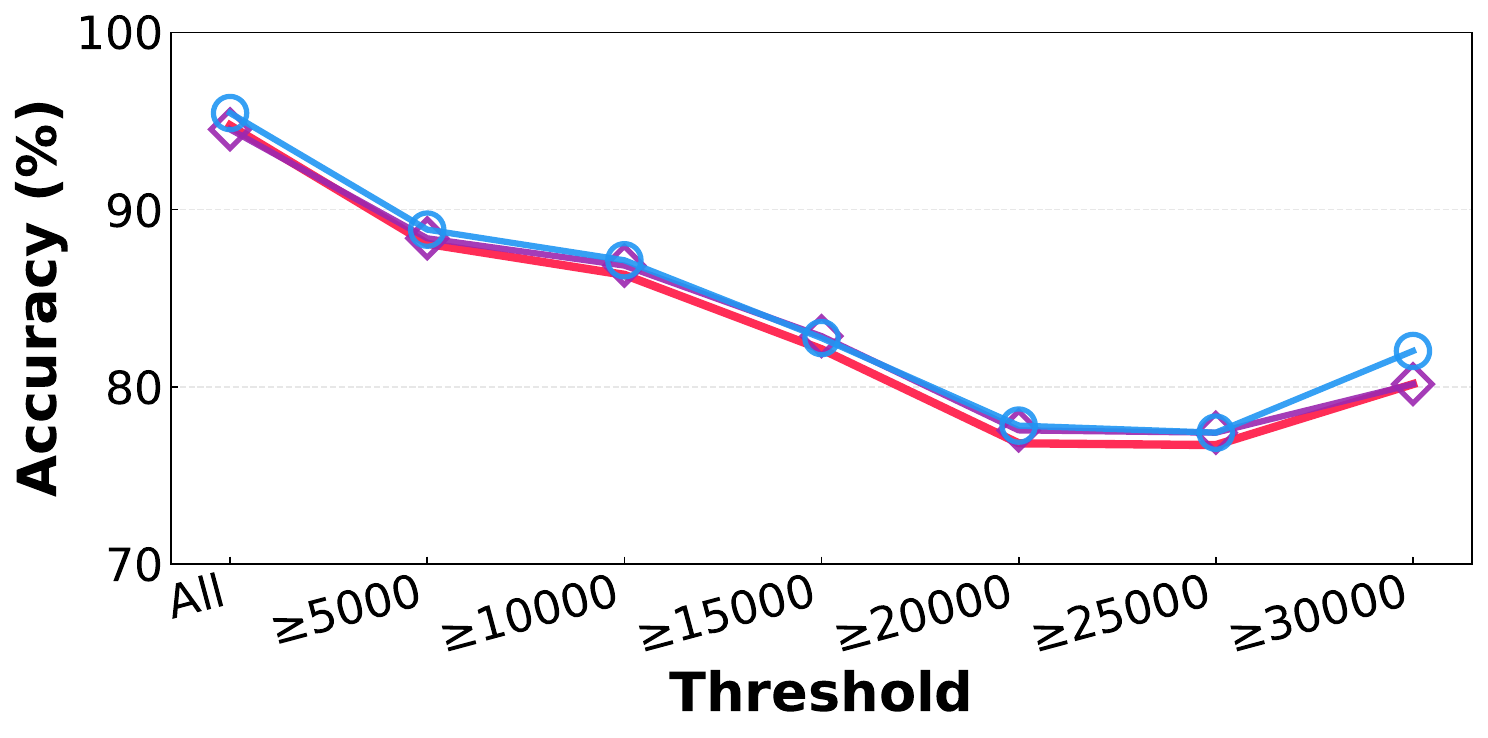}
    \subcaption{Gemini-3 Flash: Acc.}
  \end{minipage}\hfill
  \begin{minipage}[t]{0.32\linewidth}
    \centering
    \includegraphics[width=\linewidth]{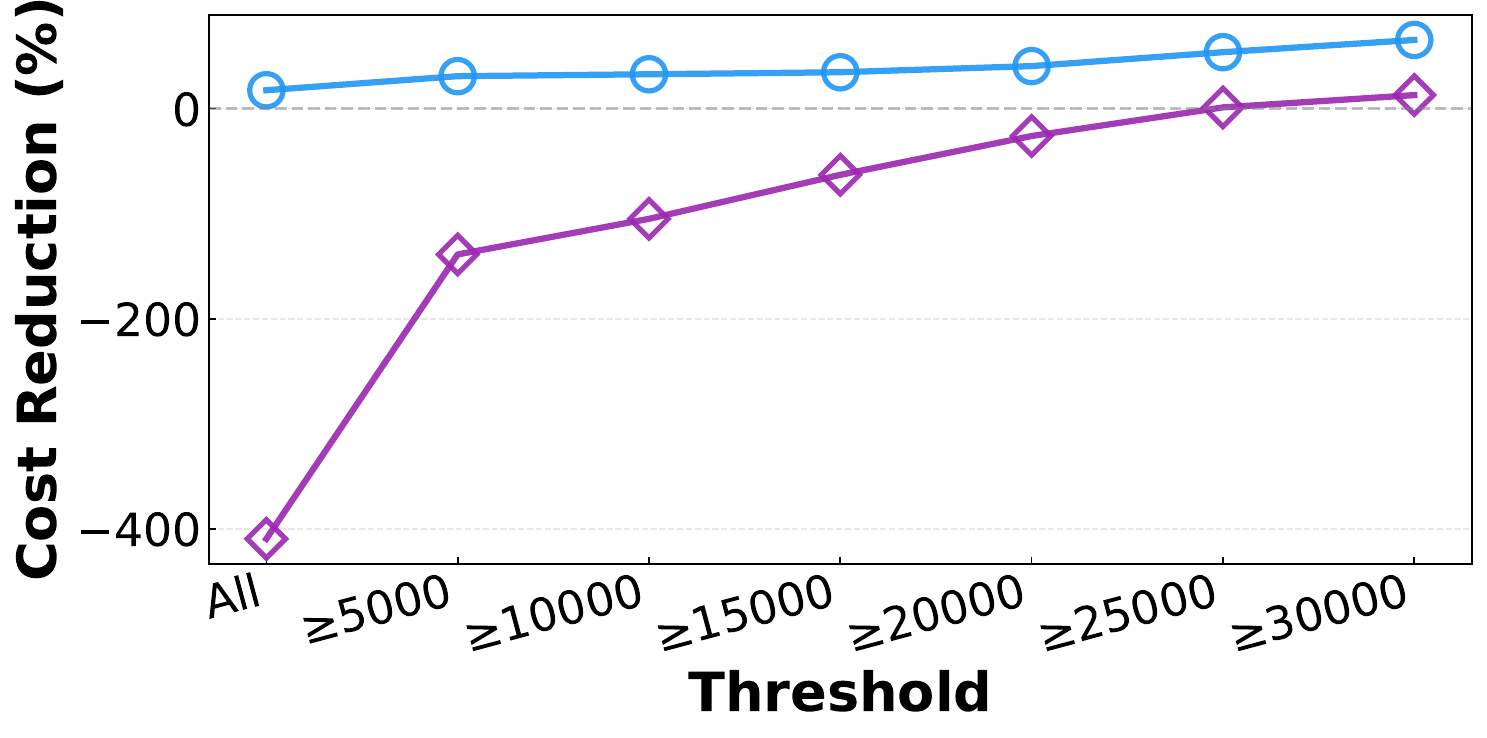}
    \subcaption{Gemini-3 Flash: Cost}
  \end{minipage}\hfill
  \begin{minipage}[t]{0.32\linewidth}
    \centering
    \includegraphics[width=\linewidth]{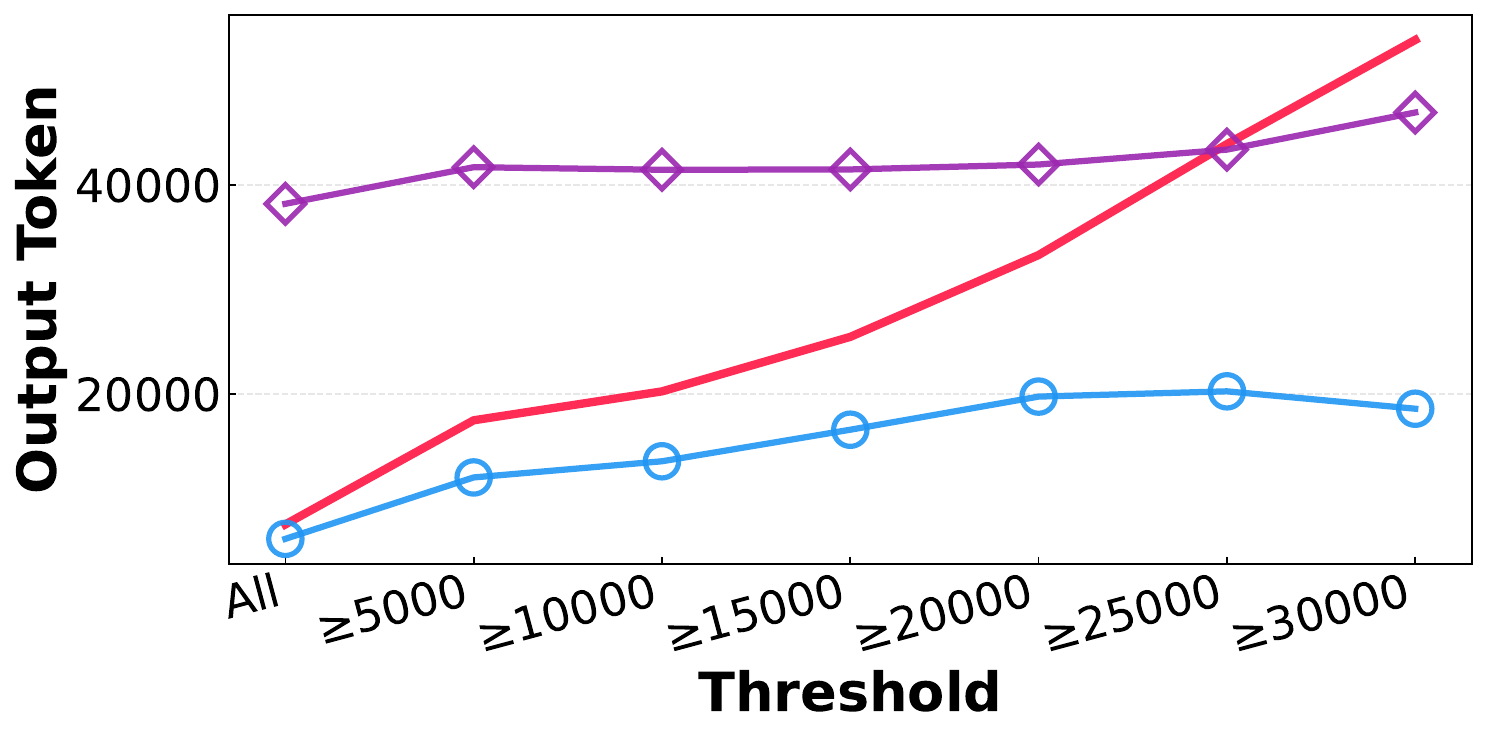}
    \subcaption{Gemini-3 Flash: Tokens}
  \end{minipage}

  \vspace{0.5em}

  \begin{minipage}[t]{0.32\linewidth}
    \centering
    \includegraphics[width=\linewidth]{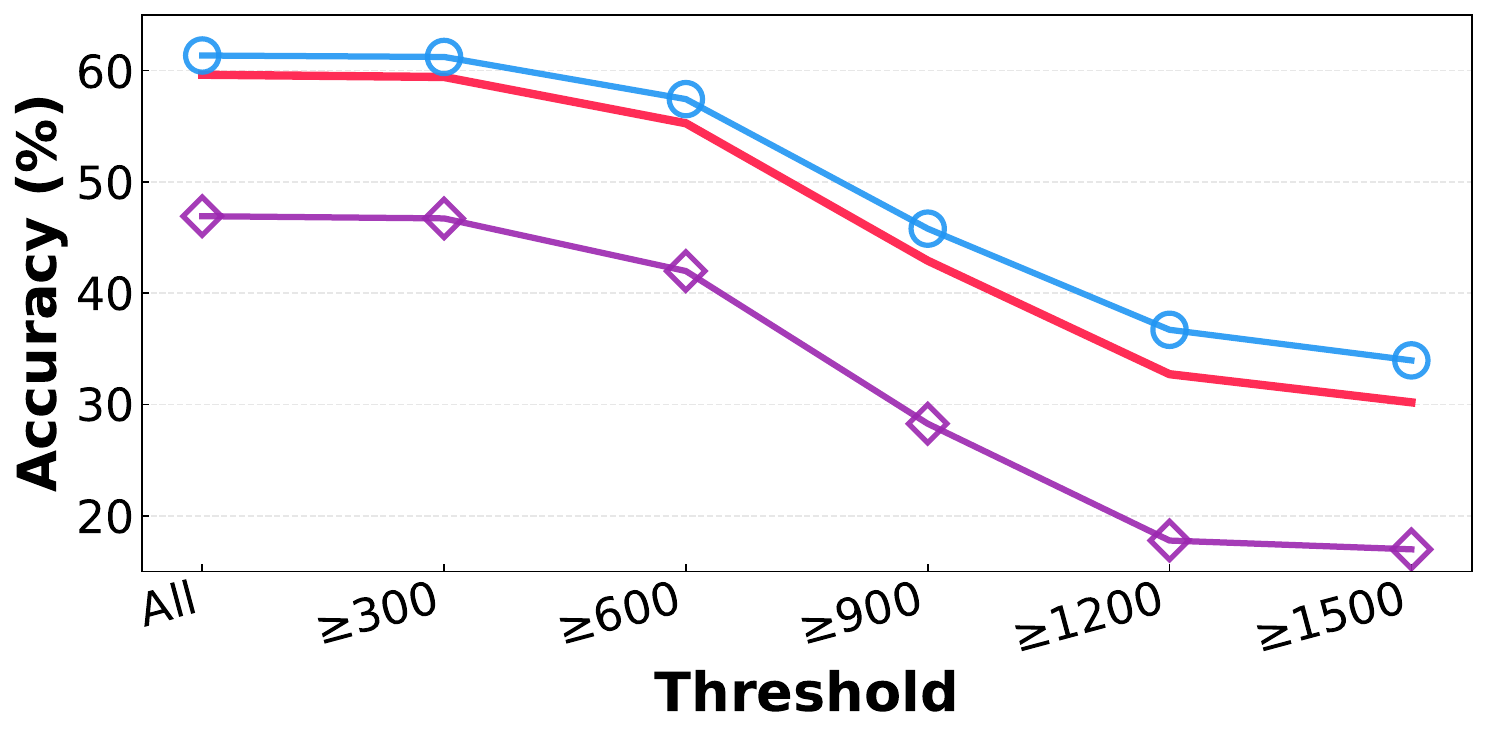}
    \subcaption{GPT-4o-mini: Acc.}
  \end{minipage}\hfill
  \begin{minipage}[t]{0.32\linewidth}
    \centering
    \includegraphics[width=\linewidth]{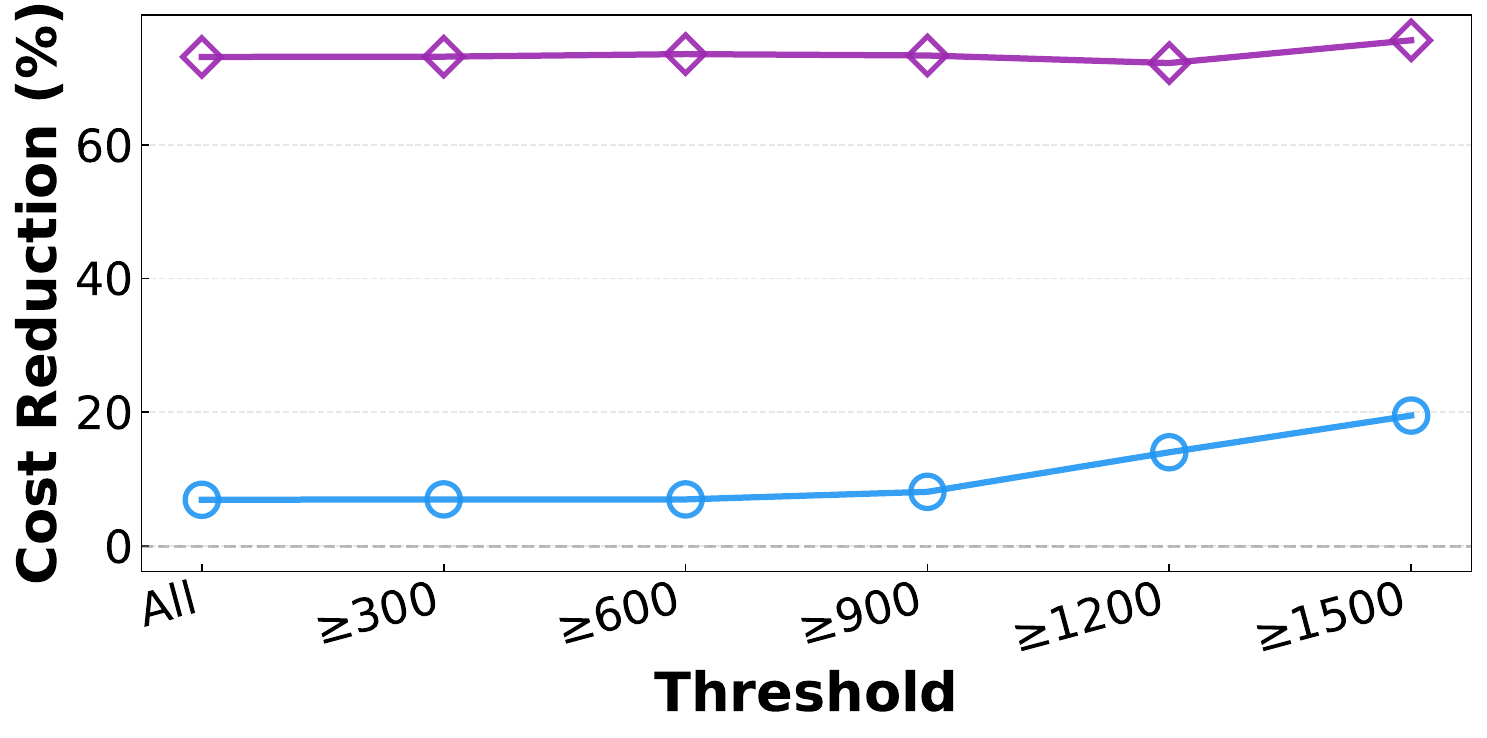}
    \subcaption{GPT-4o-mini: Cost}
  \end{minipage}\hfill
  \begin{minipage}[t]{0.32\linewidth}
    \centering
    \includegraphics[width=\linewidth]{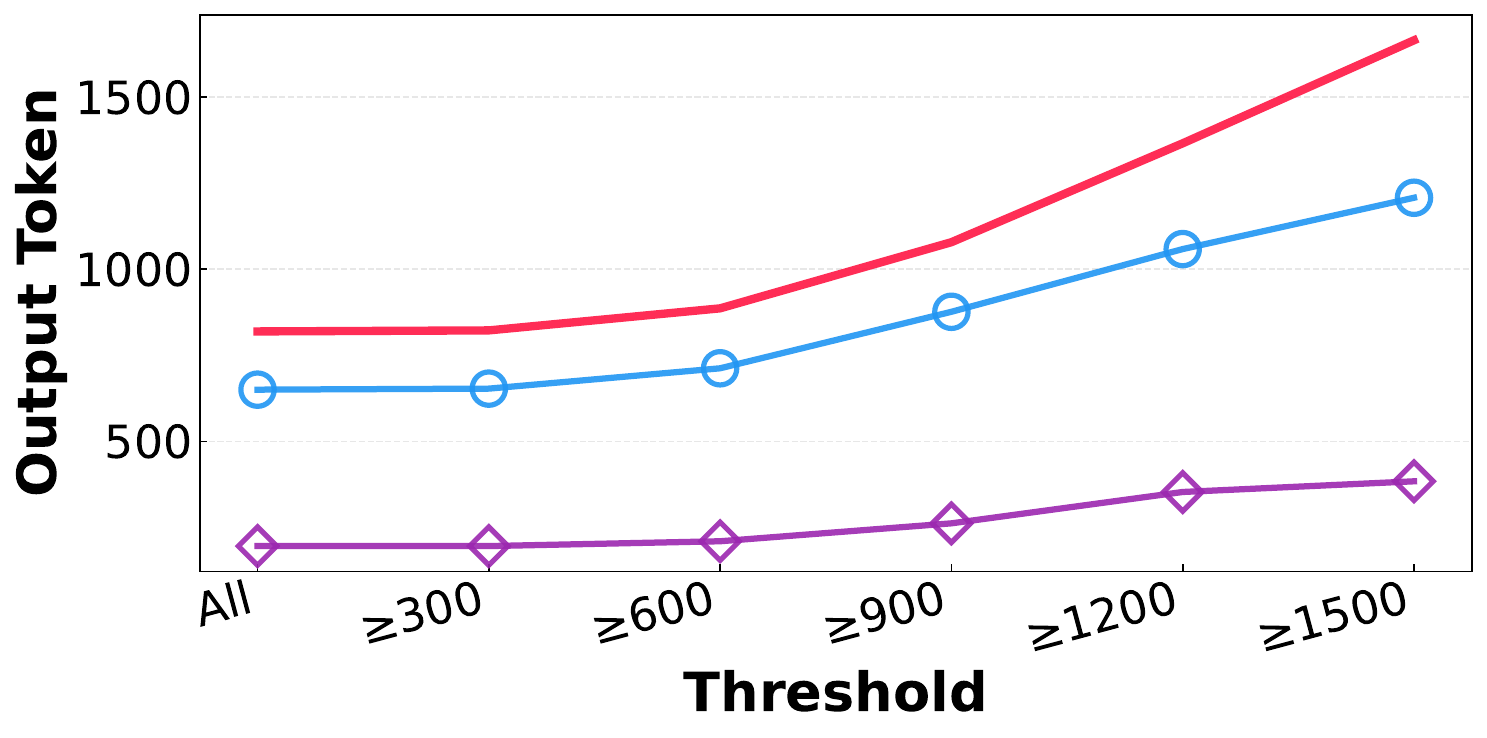}
    \subcaption{GPT-4o-mini: Tokens}
  \end{minipage}

  \caption{Main results compared with CoD at different thresholds (DeepMath-103K).}
  \label{fig:generalize-9}
  %\vspace{-0.1cm}
\end{figure*}

In this section, we extend the comparison between \textbf{Thinking with Reasoning Skills (TRS)} and the \textbf{Chain-of-Draft (CoD)} baseline to five additional models: \textbf{GPT-5.2}, \textbf{Grok-4-Fast}, \textbf{Gemini-3-Pro}, \textbf{Gemini-3-Flash}, and \textbf{GPT-4o-mini}.
This analysis (visualized in Figure~\ref{fig:generalize-9}) complements Table~\ref{tab:main} by breaking down performance across varying difficulty thresholds ($\theta$), where a higher $\theta$ indicates problems that require longer reasoning traces from the baseline model.

\begin{figure*}[h!]
\setlength{\abovecaptionskip}{0.1cm}
    \centering
    \includegraphics[width=\textwidth]{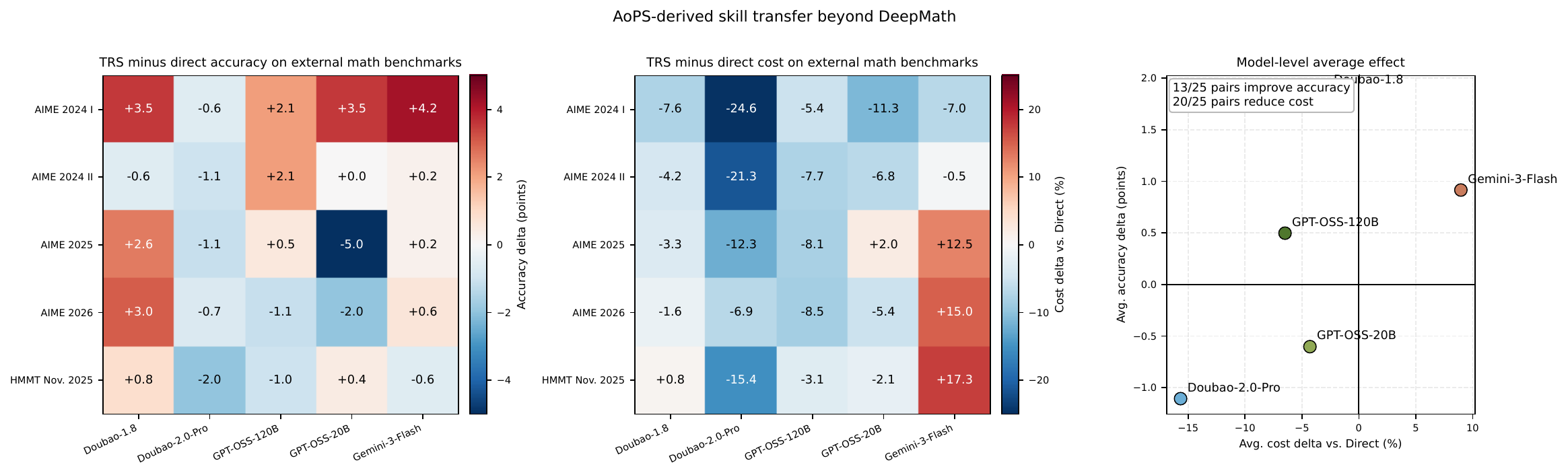}
    \caption{External contest-math transfer with the AoPS-derived skill bank. Left: TRS-minus-direct accuracy deltas. Middle: TRS-minus-direct cost deltas as percentages relative to Direct. Right: model-level average accuracy/cost-percentage trade-off. The overall pattern is mixed: 13 of 25 pairs improve accuracy and 20 of 25 reduce cost.}
    \label{fig:aops-benchmark}
\end{figure*}

\begin{table*}[t]
    \centering
    \tiny
    \setlength{\tabcolsep}{6pt}
    \resizebox{\textwidth}{!}{
    \begin{tabular}{lcccc}
        \toprule
        Benchmark & Doubao Direct $\rightarrow$ TRS & Doubao Cost $\Delta$ & OSS Direct $\rightarrow$ TRS & OSS Cost $\Delta$ \\
        \midrule
        SST-2 & $93.5 \rightarrow 92.5$ & $+93.3\%$ & $92.5 \rightarrow 92.5$ & $+72.9\%$ \\
        CoLA & $86.5 \rightarrow 85.5$ & $+102.8\%$ & $82.0 \rightarrow 84.5$ & $+52.4\%$ \\
        MRPC & $68.5 \rightarrow 67.0$ & $+43.6\%$ & $77.0 \rightarrow 79.0$ & $+36.8\%$ \\
        AG News & $88.0 \rightarrow 87.5$ & $+64.1\%$ & $82.5 \rightarrow 85.0$ & $+74.7\%$ \\
        \bottomrule
    \end{tabular}}
    \caption{Non-reasoning stress test with a deliberately mismatched DeepMath skill library. Cost deltas are percentages relative to Direct. Skill injection does not cause catastrophic degradation, but it consistently increases cost and is not uniformly beneficial on short-form classification tasks.}
    \label{tab:nonreasoning-stress}
\end{table*}

\begin{table}[t]
    \centering
    \small
    \setlength{\tabcolsep}{5pt}
    \begin{tabular}{lccc}
        \toprule
        Benchmark & Acc. $\Delta$ & Out Tok. $\Delta$ & Cost $\Delta$ \\
        \midrule
        AIME 2024 I & +2.54 & $\downarrow$11.4\% & $\downarrow$10.8\% \\
        AIME 2024 II & +0.12 & $\downarrow$7.4\% & $\downarrow$6.3\% \\
        AIME 2025 & -0.56 & $\uparrow$1.2\% & $\uparrow$4.9\% \\
        AIME 2026 & -0.04 & $\uparrow$1.9\% & $\uparrow$7.3\% \\
        HMMT Nov.\ 2025 & -0.48 & $\uparrow$3.0\% & $\uparrow$7.1\% \\
        \bottomrule
    \end{tabular}
    \caption{Benchmark-level averages on the external contest-math suite. Accuracy is reported in percentage points; output tokens and cost are relative changes against Direct. AIME 2024 I is the clearest positive transfer regime, while newer or harder sets show smaller or mixed effects.}
    \label{tab:aops-benchmark-by-benchmark}
\end{table}

The results highlight two critical advantages of TRS over CoD:
\begin{enumerate}
    \item \textbf{Robustness on Hard Problems:} As difficulty increases (moving right on the x-axis), CoD frequently exhibits a sharp decline in accuracy (e.g., see GPT-4o-mini and Gemini-3-Pro panels). This confirms that simply forcing brevity often compromises reasoning depth on complex tasks. In contrast, TRS consistently maintains or improves upon baseline accuracy, effectively breaking the efficiency-accuracy trade-off.
    \item \textbf{Consistent Cost Reduction:} While CoD achieves aggressive token reduction, it often does so at the expense of correctness. TRS achieves competitive cost reductions (e.g., significant savings on Gemini-3-Flash and GPT-5.2) while ensuring that the "thinking" process remains guided and correct via retrieved skills.
\end{enumerate}

\section{External Contest-Math Benchmark Transfer with AoPS Skills}
\label{app:aops-benchmark}

The external benchmark study uses a \emph{pure} AoPS-derived skill library consisting of 7,616 cards distilled from a broad contest-math corpus, including AHSME, AMC, AIME, USAMO, USAJMO, and IMO-style problems. The evaluation suite contains 120 questions in total: 15 from AIME 2024 I, 15 from AIME 2024 II, 30 from AIME 2025, 30 from AIME 2026, and 30 from HMMT November 2025. In this benchmark matrix, TRS builds a BM25 index over each skill card's question, topic, heuristic, and keywords, retrieves the top-1 match, and injects only the retrieved heuristic into the prompt. Figure~\ref{fig:aops-benchmark} visualizes the full set of 25 model--benchmark results, including accuracy deltas, cost-percentage deltas, and the average trade-off at the model level, while Table~\ref{tab:aops-benchmark-by-benchmark} reports the benchmark-level averages. Importantly, all results reported in this section use the 7,616-card AoPS library rather than the later 7,736-card AoPS+benchmark merged library. Therefore, the transfer setting does not retrieve hints distilled from the benchmark suite itself.

The main takeaway is that AoPS-derived reasoning skills transfer \emph{non-uniformly} across models and benchmarks. The strongest positive cells are Gemini-3-Flash on AIME 2024 I ($+4.16$), GPT-OSS-20B on AIME 2024 I ($+3.55$), and Doubao-1.8 on AIME 2024 I ($+3.54$), while the strongest negative cell is GPT-OSS-20B on AIME 2025 ($-5.00$). Because the direct and TRS settings use different repeat counts (32 versus 8), we interpret these results as descriptive evidence of transfer rather than as a formal hypothesis test. Overall, this analysis strengthens the camera-ready paper by showing that TRS is not confined to a single in-domain benchmark, although its benefits remain dependent on both target model and workload.

\begin{figure}[ht]
    \centering
    \includegraphics[width=\linewidth]{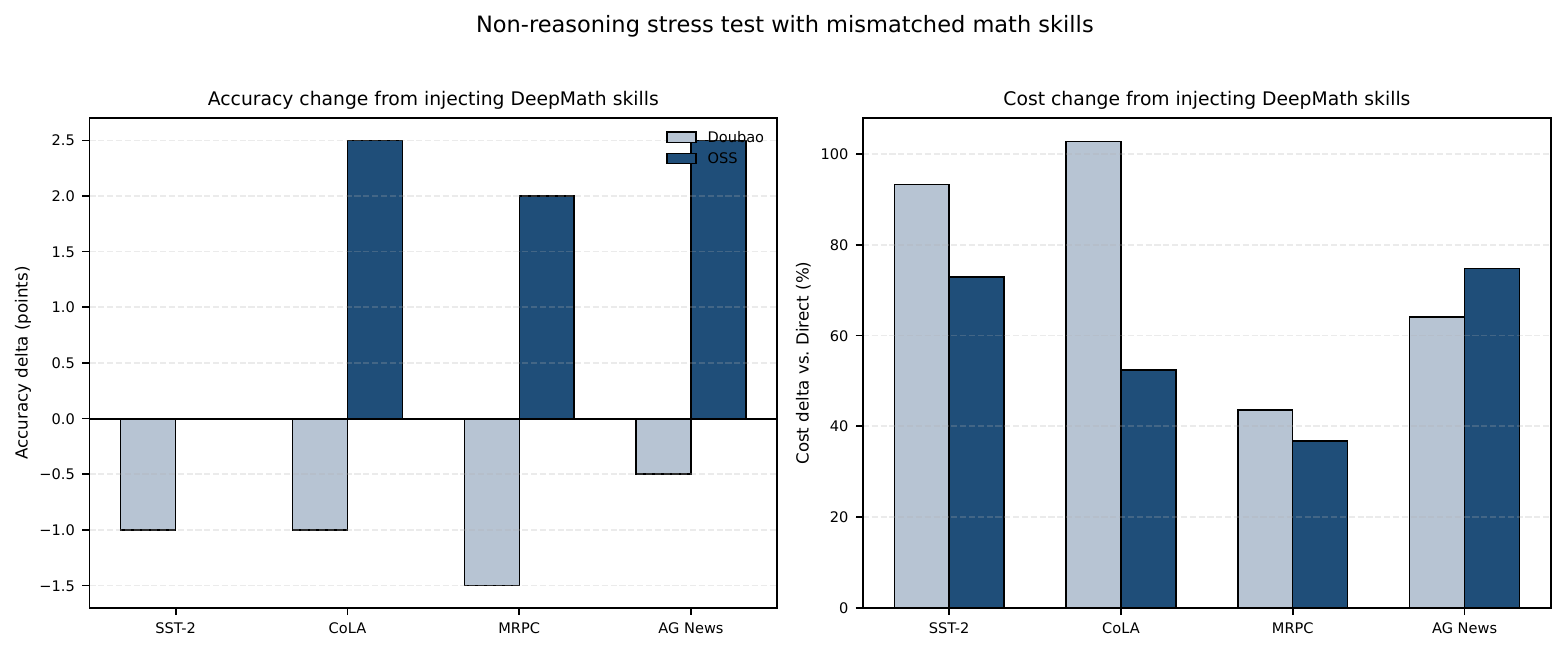}
    \caption{Accuracy and cost-percentage deltas in the non-reasoning stress test.}
    \label{fig:nonreasoning-stress}
\end{figure}

\section{Non-reasoning Stress Test}
\label{app:nonreasoning-stress}

Table~\ref{tab:nonreasoning-stress} and Figure~\ref{fig:nonreasoning-stress} summarize a deliberately mismatched stress test in which DeepMath skills are injected into short-form classification tasks. This setting is intended as a robustness control rather than a recommended deployment scenario.

Across the four non-reasoning benchmarks, the average accuracy shift is $-1.0$ point for Doubao and $+1.75$ points for OSS, while the average cost increases by $75.9\%$ and $59.2\%$, respectively. We therefore view this result primarily as a \emph{robustness control}: it suggests that injecting mismatched reasoning skills does not lead to catastrophic degradation in accuracy, but it also provides no evidence that TRS should be enabled by default for mixed or low-reasoning workloads, especially given the substantial cost overhead.

\end{document}